\documentclass[twoside,11pt]{article}
\usepackage{jmlr2e}

\usepackage{times}

\usepackage{amsmath}
\usepackage{amsfonts}
\usepackage{graphicx}
\usepackage{latexsym}
\usepackage{dsfont}
\usepackage{amssymb}
\usepackage{mathrsfs}
\usepackage{color}
\newcommand{\normop}[1]{\norm{#1}_{\op}}

\newcommand{\norm}[1]{\|#1\|}%
\newcommand{\bb}{\mathbb}

\newcommand{\ind}[1]{\mathbf 1_{#1}}%

\newcommand{\E}{\mathbb E}

\renewcommand{\P}{\mathbb P}

\newcommand{\sS}{\mathscr S}

\newcommand{\tT}{\mathds{T}}
\newcommand{\tH}{\mathds{H}}

\newcommand{\bA}{{\boldsymbol A}}%
\newcommand{\bbA}{{\mathbb{A}}}%
\newcommand{\bB}{{\boldsymbol B}}%
\newcommand{\bC}{{\boldsymbol C}}%
\newcommand{\bF}{{\boldsymbol F}}%
\newcommand{\bH}{{\boldsymbol H}}%
\newcommand{\bbH}{{\mathbb H}}%
\newcommand{\bI}{{\boldsymbol I}}%
\newcommand{\bJ}{{\boldsymbol J}}%
\newcommand{\bM}{{\boldsymbol M}}%
\newcommand{\bN}{{\boldsymbol N}}%
\newcommand{\bO}{{\boldsymbol 0}}%
\newcommand{\bP}{{\boldsymbol P}}%
\newcommand{\bU}{{\boldsymbol U}}%
\newcommand{\bV}{{\boldsymbol V}}%
\newcommand{\hbV}{{\widehat {\boldsymbol V}}}%
\newcommand{\bW}{{\boldsymbol W}}%
\newcommand{\bbW}{{\mathbb{W}}}%
\newcommand{\bX}{{\boldsymbol X}}%
\newcommand{\bZ}{{\boldsymbol Z}}%
\newcommand{\bbZ}{{\mathbb Z}}%
\newcommand{\bSigma}{{\boldsymbol \Sigma}}%
\newcommand{\blambda}{{\boldsymbol \lambda}}%

\newcommand{\lmax}{\lambda_{\max}}

\newcommand{\bul}{\bullet}

\DeclareMathOperator*{\supp}{supp}
\DeclareMathOperator*{\rank}{rank}

\DeclareMathOperator*{\argmin}{argmin}

\DeclareMathOperator{\pen}{pen}

\DeclareMathOperator{\op}{op}
\DeclareMathOperator{\tr}{tr}
\DeclareMathOperator{\sign}{sign}
\DeclareMathOperator{\diag}{diag}

\DeclareMathOperator{\hstack}{hstack}

\newcommand{\1}{{\rm 1}\kern-0.24em{\rm I}}

\newcommand{\bs}{\boldsymbol}

\newcommand{\cF}{\mathcal F}

\newcommand{\cP}{\mathcal P}

\newcommand{\R}{\mathbb R}
\newcommand{\N}{\mathbb N}

\newcommand{\inr}[1]{\langle #1 \rangle}

\newcommand{\grad}{\nabla}

\newcommand{\mleq}{\preccurlyeq}

\title{Sparse and low-rank multivariate Hawkes processes}

\jmlrheading{?}{?}{??}{??}{??}{??}{Emmanuel Bacry, Martin Bompaire, St\'ephane Ga\"iffas and Jean-Francois Muzy}

\ShortHeadings{Sparse and low-rank multivariate Hawkes processes}{Bacry, Bompaire, Ga\"iffas and Muzy}

\firstpageno{1}

\begin{document}

\title{Sparse and low-rank multivariate Hawkes processes}

\author{%
\name Emmanuel Bacry \email emmanuel.bacry@polytechnique.edu \\
\addr CEREMADE, CNRS UMR 7534, Universit\'e Paris-Dauphine, Paris, France
\AND 
\name Martin Bompaire \email m.bompaire@criteo.com \\
\addr Criteo, Paris, France
\AND 
\name St\'ephane Ga\"iffas \email stephane.gaiffas@lpsm.paris \\
\addr LPSM, CNRS UMR 8001, Universit\'e Paris-Diderot, Paris, France \\
\addr DMA, CNRS UMR 8553, Ecole Normale Supérieure, Paris, France
\AND
\name Jean-Francois Muzy \email muzy@univ-corse.fr \\
\addr Laboratoire Sciences Pour l'Environnement, CNRS UMR 6134, Universit\'e de Corse, Cort\'e, France
}

\editor{?}

\maketitle

\begin{abstract}
  We consider the problem of unveiling the implicit network structure of node interactions (such as user interactions in a social network), based only on high-frequency timestamps.
  Our inference is based on the minimization of the
  least-squares loss associated with a multivariate Hawkes model,
  penalized by $\ell_1$ and trace norm of the interaction tensor.
  We provide a first theoretical analysis for this
  problem, that includes sparsity and low-rank inducing penalizations.
  This result involves a new data-driven concentration inequality for
  matrix martingales in continuous time with observable variance, which is a result
  of independent interest and a broad range of possible applications since it extends to matrix
  martingales former results restricted to the scalar case.
  A consequence of our analysis is the construction of sharply tuned
  $\ell_1$ and trace-norm penalizations, that leads to a data-driven
  scaling of the variability of information available for each
  users.
  Numerical experiments illustrate the significant improvements
  achieved by the use of such data-driven penalizations.

\end{abstract}

\section{Introduction}

Understanding the dynamics of social interactions is a challenging
problem of rapidly growing interest \citep{PhysRevLett.92.028701,
  Leskovec08, crane2008robust, leskovec2009meme} because of the large
number of applications in web-advertisement and e-commerce, where
large-scale logs of event history are available.  A common supervised
approach consists in the prediction of labels based on declared
interactions (friendship, like, follower, etc.).  However such
supervision is not always available, and it does not always describe
accurately the level of interactions between users. Labels are often
only binary while a quantification of the interaction is more
interesting, declared interactions are often deprecated, and more
generally a supervised approach is not enough to infer the latent
communities of users, as temporal patterns of actions of users are
much more informative.

For latent social groups recovering, several recent papers~\citep{rodriguez2011uncovering,gomez13,gomez14a}
consider an approach directly
based on the real \emph{actions} or \emph{events} of users (referred to as {\em nodes} in the following) that are fully identified through their corresponding user id and timestamp. 
These models assume a structure of data
consisting in a sequence of independent cascades, containing
the timestamp of each node. In these works, techniques coming from
survival analysis are used to derive a tractable convex likelihood,
that allows one to infer the latent community structure. However, they
require that data are already segmented into sets of independent
cascades, which is often unrealistic. Moreover, it does not allow
for recurrent events, namely a node can be infected only once, and it
cannot incorporate exogenous factors, i.e., influence from the world
outside the network.

Another approach is based on self-exciting point processes, such as
the Hawkes process~\citep{hawkes1971spectra}. Previously used for
geophysics \citep{ogata1998space}, high-frequency finance
\citep{bacry2013modelling,bacryHawkesFinance}, crime activity \citep{mohler2011self}, these
processes have been recently used for the modelization of users
activity in social networks, see for instance~\cite{crane2008robust,
  blundell2012modelling, zhou2013learning, yang2013mixture}.
The structure of the Hawkes model allows us to capture the
direct influence of a specific user's action on all the
future actions of all the users (including himself). It encompasses in
a single likelihood the decay of the influence over time, the levels
of interaction between nodes, which can be seen as a weighted
asymmetrical adjacency matrix, and a baseline intensity, that measures
the level of exogeneity of a user, namely the spontaneous apparition of
an action, with no influence from other nodes of the network.

In this paper, we consider such a multivariate Hawkes process (MHP), and we
combine convex proxies for sparsity and low-rank of the adjacency
tensor and the baseline intensities, that are now of common use in
low-rank modeling in collaborative filtering problems~\citep{Candes04, Candes09}.
Note that this approach is also considered in~\citep{zhou2013learning}.
We provide a first theoretical analysis of the generalization error for this problem, 
see~\cite{hansen_reynaud_bouret_viroirard} for an analysis including only entrywise $\ell_1$ penalization.
Namely, we prove a sharp oracle inequality for our procedure, that includes sparsity and low-rank
inducing priors, see Theorem~\ref{thm:fast-oracle} in Section~\ref{sec:oracle-inequalities}.
This result involves a new
data-driven concentration inequality for matrix martingales in
continuous time, see Theorems~\ref{thm:concentration_counting_emp}
and~\ref{thm:concentration_counting_emp1} in Section~\ref{sec:concentration-inequalities},
that are results of independent interest, that extends previous non-commutative versions
of concentration inequalities for martingales in discrete time, see~\cite{tropp2012user}.
A consequence of our analysis is the construction of sharply tuned $\ell_1$ and trace-norm penalizations,
that leads to a data-driven scaling of the variability of information
available for each node.
We give empirical evidence of the
improvements of our data-driven penalizations, by conducting in
Section~\ref{sec:numerical_experiment} numerical experiments on
simulated data.
Since the objectives involved are convex with a smooth component, our algorithms build upon
standard batch proximal gradient descent algorithms.

\section{The multivariate Hawkes model and the least-squares functional}

Consider a finite network with $d$ nodes (each node corresponding to a user in a social network for instance).
For each node $j \in \{ 1, \ldots, d \}$, we observe the timestamps $\{ t_{j, 1},
t_{j, 2}, \ldots \}$ of actions of node $j$ on the network (a message, a click, etc.).
With each node $j$ is associated a counting process $N_j(t) = \sum_{i \geq 1}
\ind{t_{j, i} \leq t}$ and we consider the $d$-dimensional counting
process $N_t = [N_1(t) \; \cdots \; N_d(t)]^\top$, for $t \geq 0$.
We observe this process for $t \in [0, T]$.
Each $N_j$ has an intensity $\lambda_j$, meaning that
\begin{equation*}
  \P\big( N_j \text { has a jump in } [t, t + dt] \; | \; \cF_t\big)
  = \lambda_j(t) dt, \quad j = 1, \ldots, d,
\end{equation*}
where $\cF_t$ is the $\sigma$-field generated by $N$ up to time $t$.
The multivariate Hawkes model assumes that each $N_j$ has an intensity $\lambda_{j, \theta}$ given by
\begin{equation}
  \label{eq:general_intensity}
  \lambda_{j, \theta}(t) = \mu_j + \sum_{j' = 1}^d \int_{(0, t)}  \varphi_{j, j'}(t - s) d N_{j'}(s),
\end{equation}
where $\mu_j \geq 0$ is the baseline intensity of $j$ (i.e., the intensity of exogenous events of node $j$) and where the functions $\varphi_{j, j'} : \R^+ \rightarrow \R$ for $j=1, \ldots, d$, called \emph{kernels}, allow to quantify the impact of node $j'$ on node $j$.
Note that the integral used in Equation~\eqref{eq:general_intensity} is a Stieljes integral, namely it simply stands for
\begin{equation*}
  \int_{(0, t)} \varphi(t - s) d N_{j'}(s) = \sum_{i \; : \; t_{j', i} \in [0,
  t)} \varphi(t - t_{j', i}).
\end{equation*}
In the paper, we consider general kernel functions $\varphi_{j,j'}(t)$ that can be written as:
\begin{equation}
  \label{eq:general_kernel}
  \varphi_{j, j'}(t) = \sum_{k=1}^K a_{j, j', k} h_{j, j', k}(t).
\end{equation}
where the coefficients $a_{j, j', k}$ are the entries of a $d \times d \times K$ tensor $\bbA$ (i.e., $(\bbA)_{j, j', k} = a_{j, j', k}$) and the kernels $h_{j, j', k}(t)$ are elements of a fixed dictionnary of non negative and causal functions ($h_{j, j', k}: \R^+ \rightarrow \R^+$) such that $\| h_{j, j', k} \|_1 = 1$. In that respect, the weights $a_{j, j', 1}, \ldots, a_{j, j', K}$ all quantify the influence of $j'$ on $j$, but the particular weight $a_{j, j', k}$ quantifies it for the $k$-th \emph{decay function} $h_{j, j', k}$.
%Indeed, each $h_{j, j', k} : \R^+ \rightarrow \R^+$ is a so-called decay function (with fixed $L^1$ norm
%$\| h_{j, j', k} \|_1 = 1$) that accounts for the decay of influence between pairs of nodes in the network.
%In order to obtain more flexibility in the modelization of the decays, we consider a fixed set of $K$ ``simple'' functions $h_{j, j', 1}, \ldots, h_{j, j', K}$ to approximate the decay of the influence of $j'$ on $j$.
A standard choice is a dictionnary of exponential kernels, $h_{j, j', k}(t) = \alpha_k e^{-\alpha_k t}$ with varying memory parameters $\alpha_1, \ldots, \alpha_K$.
This leads to the following standard parametrization of the kernel functions, called \emph{exponential kernels}:
\begin{equation}
  \label{eq:sum_of_exponentials_kernel}
  \varphi_{j, j'}(t) = \sum_{k=1}^K a_{j, j', k} \alpha_k \exp(-\alpha_k t).
\end{equation}
The main advantage of exponential kernels with fixed memory parameters $\alpha_1, \ldots, \alpha_K$,
is that it allows one to handle a convex problem.
In the general case or when the memory parameters are unknown, the problem becomes non-convex, more challenging and is beyond the scope of the paper.

The parameter of interest is the \emph{self-excitement} tensor $\bbA$, which can be viewed as a 
cross-scale (for $k=1, \ldots, K$) weighted adjacency matrix of connectivity between nodes, as illustrated in Figure~\ref{fig:first_illustration} below.

\begin{figure}[h]
  \centering
  \includegraphics[width=0.85\textwidth]{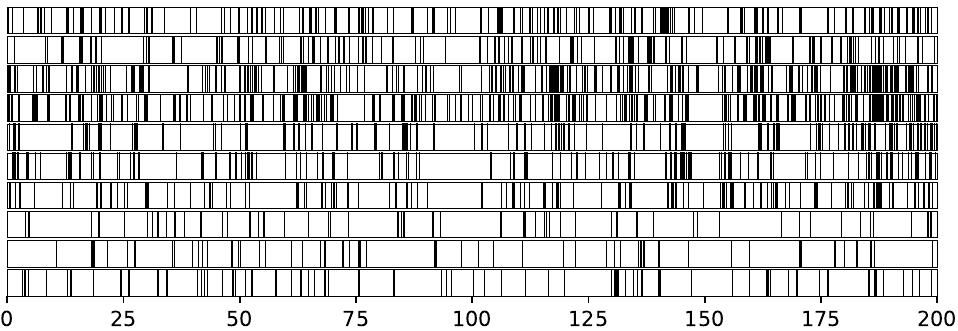}
  \includegraphics[width=0.85\textwidth]{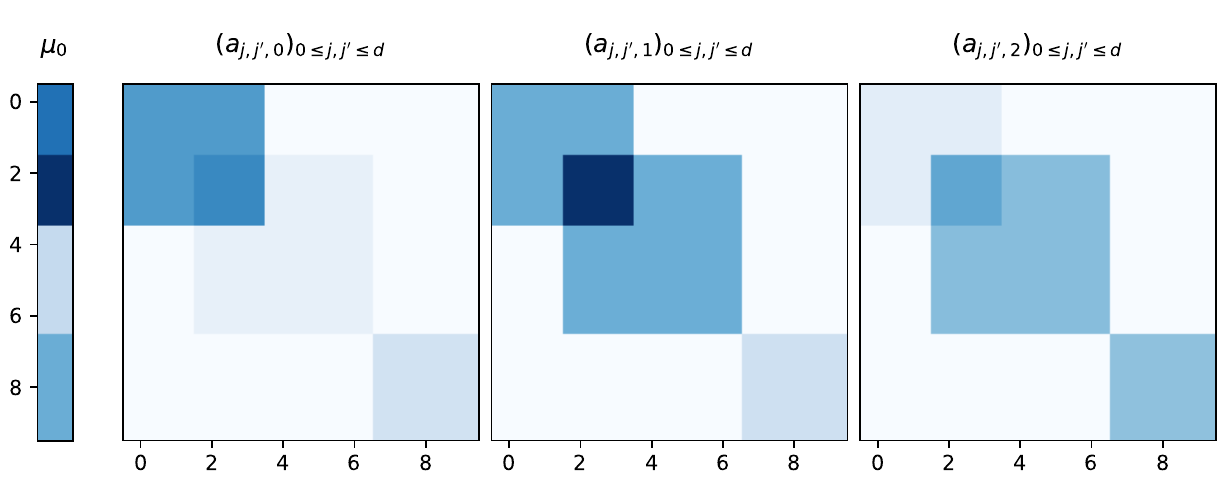}
  \caption{Toy example with $d = 10$ nodes. Based on actions' timestamps of the nodes, represented
  by vertical bars (top), we aim at recovering the vector $\mu_0$ and the tensor $\bbA$ of
  implicit influence between nodes (bottom).}
  \label{fig:first_illustration}
\end{figure}

The Hawkes model is particularly relevant for the modelization of the ``microscopic'' activity of social networks and has attracted a lot of interest in the recent literature (see~\cite{crane2008robust, blundell2012modelling, zhou2013learning, yang2013mixture,linderman2014discovering, dubois2013stochastic, blundell2012modelling, iwata2013discovering}, among others) for this kind of application, with a particular emphasis on~\cite{hansen_reynaud_bouret_viroirard} that gives first theoretical results for the Lasso used with Hawkes processes with an application to neurobiology. The main point is that this simple autoregressive structure of the intensity allows us to capture the direct influence of a user, based on the recurrence and the patterns of his actions, by separating the intensity into a baseline and a self-exciting component, hence allowing to filter exogeneity in the estimation of users' influences on each others.

We introduce in this paper an estimation procedure of $\theta = (\mu, \bbA)$ based on data $\{ N_t : t \in [0, T ]\}$.
The hidden structure underlying the observed actions of nodes is contained in $\bbA$.
Our strategy is based on the least-squares functional given by
\begin{equation}
  \label{eq:least-squares}
  R_T(\theta) = \norm{\lambda_\theta}_T^2 - \frac2T \sum_{j=1}^d
  \int_{[0, T]} \lambda_{j, \theta}(t) d N_j(t),
\end{equation}
with respect to $\theta$, where
$\norm{\lambda_\theta}_T^2 = \frac 1T \sum_{j = 1}^d \int_{[0, T]}
\lambda_{j, \theta}(t)^2 dt$ is the norm associated with the inner
product
\begin{equation}
  \label{eq:inner-product}
  \inr{\lambda_\theta, \lambda_{\theta'}}_T = \frac 1T \sum_{j = 1}^d
  \int_{[0, T]} \lambda_{j, \theta}(t) \lambda_{j, \theta'}(t) dt.
\end{equation}
This least-squares function is very natural, and comes from the empirical risk minimization principle~\citep{van2000empirical, massart2007concentration, koltchinskii2011oracle,bartlett2006empirical}: assuming that $N_j$ has an unknown ground truth intensity $\lambda_j$ (not necessarily following the Hawkes model), the Doob-Meyer's decomposition gives
\begin{equation*}
  \int_{[0, T]} \lambda_{j, \theta}(t) d N_j(t) = \int_{[0, T]}
  \lambda_{j, \theta}(t) \lambda_j(t) d t + \int_{[0, T]}
  \lambda_{j, \theta}(t) d M_j(t),
\end{equation*}
where $M_j(t) = N_j(t) - \int_0^t \lambda_j(s) ds$ is a continuous-time martingale with upwards jumps of +1.
Since the ``noise'' term $\int_{[0, T]} \lambda_{j, \theta}(t) d M_j(t)$ is centered, we obtain
\begin{equation*}
  \E[ R_T(\theta) ] = \E \norm{\lambda_\theta}_T^2 - 2 \E\inr{\lambda_\theta, \lambda}_T = \E \norm{\lambda_\theta - \lambda}_T^2 - \norm{\lambda}_T^2,
\end{equation*}
so that we expect a minimum $\hat \theta$ of $R_T(\theta)$ to lead to a good estimation $\lambda_{\hat \theta}$ of $\lambda$, following the empirical risk minimization principle.
As explained in Section~\ref{sec:proofs} below, the noise terms can be written as
\begin{equation*}
    \int_0^t \tT_s \circ  d \bM_s,
\end{equation*}
for a specific tensor $\tT_t$ and matrix martingale $\bM_t$, where $\tT_s \circ \bM_s$ stands for a tensor-matrix product defined in Section~\ref{sub:notation} below.
The next Section introduces new results, of independent interest, providing \emph{data-driven} deviation inequalities for the operator norm of a matrix martingale defined as the stochastic integral $\int_0^t \tT_s \circ  d \bM_s$.
These results allow us, as a by-product, to control the noise terms arising in the application considered in this paper, and lead to a sharp data-driven tuning of the penalizations used on $\bbA$, as explained in Section~\ref{sec:procedure} below.

\section{A new data-driven matrix martingale Bernstein's inequality}

An important ingredient for the theoretical results proposed in this paper is an observable deviation inequality for continuous time matrix martingales.
We first recall previous results obtained in~\cite{bacry_gaiffas_muzy_1} about non-observable deviation inequalities for such objects.

\subsection{Notations}
\label{sub:notation}

Let $\tT$ be a tensor of shape $m \times n \times p \times q$. 
It can be considered as a linear mapping from
$\R^{p \times q}$ to $\R^{m \times n}$ according to the following ``tensor-matrix'' product:
\begin{equation*}
(\tT \circ \bA )_{i,j} = \sum_{k=1}^p \sum_{l=1}^q \tT_{i,j;k,l} \bA_{k,l}.
\end{equation*}
We will denote by $\tT^{\top}$ the tensor such that $\tT^\top \circ \bA = (\tT \circ \bA)^{\top}$ (i.e., $\tT^{\top}_{i,j;k,l} = \tT_{j,i;k,l}$)
and by $\tT_{\bul, \bul;k,l}$ and $\tT_{i,j;\bul, \bul}$ the matrices obtained when fixing the indices $k,l$ and $i,j$ respectively.
Note that $(\tT \circ \bA )_{i,j} = \tr (\tT_{i,j;\bul, \bul} \bA^{\top})$.
If $\tT$ and $\tT'$ are two tensors of dimensions $m \times n \times p \times q$ and $n \times r \times p \times q$ respectively, $\tT \tT'$ stands for the $m \times r \times p \times q$ tensor
 defined as $(\tT \tT')_{i,j;k,l} = (\tT_{\bul, \bul;k,l} \tT'_{\bul, \bul;k,l})_{i,j}$.
Accordingly, for an integer $r \geq 1$, if $\tT_{\bul,\bul;a,b}$ are square matrices, we will denote by $\tT^{r}$ the tensor such that
$(\tT^{r})_{i,j;k,l} = (\tT_{\bul, \bul;k,l}^r)_{i,j}$.
We also introduce $\norm{\tT}_{\op; \infty} = \max_{k, l}
\normop{\tT_{\bul, \bul; k, l}}$, the maximum operator norm of all matrices formed by the first two dimensions of tensor $\tT$.

In this paper we shall consider the class of $m \times n$ matrix martingales that can be written as
\begin{equation}
\label{eq:martingale}
\bZ_\tT(t) = \int_0^t \tT_s \circ  d \bM_s,
\end{equation}
where $\tT_s$ is a tensor with dimensions $m \times n \times p \times q$, whose components are assumed to be locally bounded predictable random functions.
The process $\bM_t$ is a $p \times q$ is matrix with entries that are square integrable martingales with a diagonal quadratic covariation matrix.
More explicitly, the entries of $\bZ_\tT(t)$ are given by
\begin{equation*}
(\bZ_\tT(t))_{i, j} = \sum_{k=1}^p \sum_{l=1}^q  \int_0^t (\tT_s)_{i,j;k,l}  (d \bM_s)_{k,l},
\end{equation*}
where the martingale $\bM_t$ is a matrix of compensated counting processes
$\bM_t = \bN_t - \blambda_t$ where $\bN_t$ is a $p \times q$ matrix counting process  \textup(i.e., each component is a counting process\textup) with an intensity process $\blambda_t$ which is predictable, continuous and with finite variations \textup(FV\textup).

\subsection{A non-observable matrix martingale Bernstein's inequality}

The next Theorem (which is a small variation of Theorem 2 in~\cite{bacry_gaiffas_muzy_1}) provides
a concentration inequality for $\normop{\bZ_\tT(t)}$, the operator norm of $\bZ_\tT(t)$.
Before stating the Theorem, let us introduce some more notations.
We define
\begin{equation}
\label{eq:b_t_def}
b_\tT(t) = \sup_{0 \leq s \leq t} \max \big( \norm{\tT_s}_{\op; \infty},
\norm{\tT_s^\top}_{\op; \infty} \big),
\end{equation}
and depending on whether the tensor $\tT_s$ is symmetric (i.e., $\tT_s^\top = \tT_s$ and $m = n$) or not, we define the following.
\begin{itemize}
\item If $\tT_s$ is symmetric, we define
\begin{equation}
\label{eq:W_def2}
\bW_{\tT}(s) = \tT_s^2 \circ \blambda_s
\end{equation}
and $K_{m,n} = m$
\item If $\tT_s$ is not symmetric, we define
\begin{equation}
\label{eq:W_def1}
\bW_{\tT}(s) =
\begin{bmatrix}
\tT_s \tT_s^\top \circ \blambda_s  & \bO \\ \bO &
\tT_s^\top \tT_s \circ  \blambda_s
\end{bmatrix},
\end{equation}
and $K_{m,n} = m + n$.
\end{itemize}
In both cases, we define
\begin{equation}
\label{eq:def_V_t_counting}
\bV_\tT(t) = \int_0^t \bW_\tT(s) \; ds.
\end{equation}
Finally, all along the paper we denote $\phi(x) = e^{x}-1-x$ for $x \in \R$.
The following concentration inequality is an easy consequence of Theorem 1
from~\cite{bacry_gaiffas_muzy_1}.
\begin{theorem}
  \label{thm:concentration_counting_th}
  Let $\bZ_\tT(t)$ be the $m \times n$ matrix martingale given by Equation~\eqref{eq:martingale}. Moreover, assume that
  \begin{equation}
  \label{eq:ass_thm1}
  \E \bigg[ \int_0^t
  \frac{\phi\big(3 \max(\norm{\tT_s}_{\op; \infty},\norm{\tT_s^\top}_{\op; \infty})\big)}{
      \max(\norm{\tT_s}^2_{\op; \infty},\norm{\tT_s^\top}^2_{\op; \infty})}
  (\bW_\tT(s))_{i, j} ds \bigg] < +\infty,
  \end{equation}
  for any $1 \leq i, j \leq m + n$. Then for any $\xi \in (0,3)$, $t, b, x > 0$, the following holds\textup:
  \begin{equation}
  \label{eqthm1:1}
  \P \bigg[ \normop{\bZ_\tT(t)} \geq  \frac{\phi(\xi)}{\xi b} \lmax \big(\bV_\tT(t))  + \frac{x b}{\xi}, \quad b_\tT(t) \leq b \bigg] \leq K_{m,n} e^{-x}.
 \end{equation}
 Optimizing this last inequality on $\xi$ gives
\begin{equation}
\label{eqcor1:1}
\P \bigg[ \normop{\bZ_\tT(t)} \geq \sqrt{2 v x } +
\frac{b x}{3}, \lmax(\bV_\tT(t)) \leq v, \quad b_\tT(t) \leq b \bigg] \leq K_{m,n} e^{-x}.
\end{equation}
\end{theorem}

The proof of Theorem~\ref{thm:concentration_counting_th} is given in Section~\ref{app:conc1} below.
This result is a Freedman (or Bernstein) inequality for the operator norm of $\bZ_\tT(t)$, that provides a deviation based on a variance term $\bV_\tT(t)$ and a $L^\infty$ term $b_\tT(t)$.
It is a strong generalization of the scalar Freedman inequality for continuous time martingales, and this result match exactly the scalar case whenever $\bZ_\tT(t)$ is scalar.
A more thorough discussion about the consequences of this result is provided in~\cite{bacry_gaiffas_muzy_1}.

\subsection{Data-driven matrix martingale Bernstein's inequalities}
\label{sec:concentration-inequalities}

Inequality~\eqref{eqcor1:1} is of poor practical interest in situations where
one observes only the jumping times of the $\bZ_t$ components (namely $\bN_t$) and not the stochastic intensity $\blambda_t$.
In that respect, one needs a "data driven" inequality where $\bV_\tT(t)$ is replaced by its empirical version $\hbV_\tT(t)$.
\begin{itemize}
\item If $\tT_s$ is symmetric, we define
\begin{equation*}
\hbV_\tT(t) =\int_0^t \tT_s^2 \circ d \bN_s,
\end{equation*}
\item while if $\tT_s$ is not symmetric, we define
\begin{equation*}
\hbV_\tT(t) =
\begin{bmatrix}
\int_0^t \tT_s \tT_s^\top \circ d \bN_s & \bO \\ \bO &
\int_0^t \tT_s^\top \tT_s \circ d \bN_s
\end{bmatrix}.
\end{equation*}
\end{itemize}
The next Proposition allows us to control $\lambda_{\max}(\bV_\tT(t))$ using its observable counterpart $\lambda_{\max}(\hbV_\tT(t))$ with a large probability.
This result is a generalization to arbitrary matrices of dimensions $m \times n$ of an analog
inequality originally proven by~\cite{hansen_reynaud_bouret_viroirard} for scalar martingales.
\begin{proposition}
	\label{lem1}
	For any $x,b>0$ and $\xi \in (0,3)$ such that $\xi > \phi(\xi)$, we have
	\begin{equation*}
	  \P \bigg[ \lambda_{\max}(\bV_\tT(t)) \geq \frac{\xi}{\xi-\phi(\xi)}
	  \lambda_{\max}(\hbV_\tT(t)) + \frac{xb^2}{\xi-\phi(\xi)}, \quad b_\tT(t) \leq
	   b \bigg] \leq K_{m,n} e^{-x},
	\end{equation*}
	where $K_{m,n}$ is defined as in Theorem \ref{thm:concentration_counting_th}.
	Moreover, choosing $\xi = -W_{-1}(-\frac{2}{3} e^{-2/3})-2/3$ \textup(note that $\xi \approx 0.762$\textup), where $W_{-1}$ is the second branch of the Lambert $W$ function, leads to
	\begin{equation*}
	\P \Big[ \lambda_{\max}(\bV_\tT(t)) \geq 2 \lambda_{\max}(\hbV_\tT(t)) + c b^2 x,
	\quad b_\tT(t) \leq b \Big] \leq K_{m,n} e^{-x}
	\end{equation*}
	for any $x,b>0$, with $c = 2.62$.
\end{proposition}
Thanks to Proposition~\ref{lem1}, we can establish an analog of Theorem 1 where
$\lambda_{\max}(\bV_\tT(t))$ is replaced by its data-driven version
$\lambda_{\max}(\hbV_\tT(t))$, up to a slight loss in values of the numerical constants.
\begin{theorem}
\label{thm:concentration_counting_emp}
With the same notations and assumptions as in Theorem~\ref{thm:concentration_counting_th} one
has
\begin{equation}
	\label{eqcor2:1}
	\P \bigg[ \normop{\bZ_\tT(t)} \geq 2 \sqrt{v x} + c b x, \;\;
	\lambda_{\max}(\hbV_\tT(t)) \leq v, \; \; b_\tT(t) \leq b \bigg]
	\leq 2 K_{m,n} e^{-x}
\end{equation}
for any $x,b >0$ with $c = 14.39$.
\end{theorem}

The proof of Theorem~\ref{thm:concentration_counting_emp} is given in Section~\ref{sub:proof_of_theorem_thm:concentration_counting_emp} below.
It follows simple arguments that combine Theorem~\ref{thm:concentration_counting_th} and Proposition~\ref{lem1}.
However, this inequality is stated on the events $\{\lambda_{\max}(\hbV_\tT(t)) \leq v\}$ and $\{ b_\tT(t) \leq b \}$, while an unconditional deviation inequality is more practical.
Such a result, which involves some extra technicalities, is stated in the next Theorem.
\begin{theorem}
\label{thm:concentration_counting_emp1}
With the same conditions and notations as in Theorem \ref{thm:concentration_counting_emp},
one has
\begin{equation}
\label{eqth3:1}
\P \bigg[ \normop{\bZ_\tT(t)} \geq 2 \sqrt{\lambda_{\max}(\hbV_\tT(t)) (x+\ell_x(t))} +
c (x+\ell_x(t)) (1 + b_\tT(t)) \bigg] \leq C_{m,n} e^{-x}
\end{equation}
where $C_{m,n} = \frac{\pi^4}{18 \log(2)^4} K_{m,n} \leq 23.45 K_{m, n}$, where $c = 14.39$ and
\begin{equation*}
	\ell_x(t) = 2 \log \log \Big (\frac{4 \lambda_{\max}(\hbV_\tT(t))}{x}
	\vee 2 \Big) + 2 \log \log (4b_\tT(t) \vee 2).
\end{equation*}
\end{theorem}
The proof of this Theorem is given in Section \ref{app:th3}.
It is a result of independent interest, that gives a control on the operator norm of a matrix martingale in continuous time (with jumps at most $1$), using only observable quantities. Along with~\cite{bacry_gaiffas_muzy_1}, it provides a first deviation inequality for such objects, and it can be understood as a data-driven version of the results given in~\cite{bacry_gaiffas_muzy_1}.

\section{The procedure}
\label{sec:procedure}

We want to produce an estimation procedure of $\theta = (\mu, \bbA)$ based on data from $\{ N_t : t \in [0, T ]\}$.
Following the empirical risk minimization principle, the estimation procedure uses the least-squares functional~\eqref{eq:least-squares} as a goodness-of-fit.
In addition to this goodness-of-fit criterion, we need to use a penalization that allows us to reduce the dimensionality of the model, namely we consider
\begin{equation}
  \label{eq:hat_theta_def_L2_risk}
  \hat \theta \in \argmin_{\theta = (\mu, \bbA) \in \R_+^d \times
  \R_+^{d \times d \times K}}
  \big\{ R_T(\theta) + \pen(\theta) \big\},
\end{equation}
for a specific penalization function $\pen(\theta)$ described below.
In particular, we want to reduce the dimensionality of $\bbA$, based on the prior assumption that latent factors explain the connectivity of users in the network.
This leads to a low-rank assumption on $\bbA$, which is commonly used in collaborative filtering and matrix completion techniques~\citep{ricci2011introduction}.
Our prior assumptions on $\mu$ and $\bbA$ are the following.

\paragraph*{Sparsity of $\mu$.}

Some nodes are basically inactive and react only if stimulated. Hence,
we assume that the baseline intensity vector $\mu$ is sparse.

\paragraph*{Sparsity of $\bbA$.}

A node interacts only with a fraction of other nodes, meaning that for
a fixed node $j$, only a few $a_{j, j', k}$ are non-zero.
Moreover, a node might react at specific time scales only, namely $a_{j, j', k}$ is non-zero for some $k$ only for fixed $j, j'$.
Hence, we assume that $\bbA$ is an entrywise sparse tensor.

\paragraph*{Low-rank of $\bbA$.}

Using together Equations~\eqref{eq:general_intensity} and~\eqref{eq:general_kernel}, one can write
\begin{equation}
  \label{eq:intensity_hstack}
  \begin{split}
    \lambda_{j, \theta}(t) &= \mu_j + \sum_{j' = 1}^d \sum_{k=1}^K a_{j, j', k} \int_{(0, t)} 
    h_{j, j', k}(t - s) d N_{j'}(s) \\
    &= \mu_j + \big(\hstack(\bbA)_{j, \bul})^\top \hstack(\bbH(t))_{j, \bul},
  \end{split}
\end{equation}
where $\bbH(t)$ is the $d \times d \times K$ tensor with entries
\begin{equation}
  \label{eq:def_bbH}
  \bbH_{j, j', k}(t) = \int_{(0, t)} h_{j, j', k}(t - s) d N_{j'}(s),
\end{equation}
where $(\bX)_{j, \bul}$ stands for the $j$-th row of a matrix $\bX$ and where $\hstack$ stands for the horizontally stacking operator defined by
\begin{equation}
  \label{eq:hstack_definition}
  \hstack : \R^{d \times d \times K} \rightarrow \R^{d \times K d} \quad \text{ such that } \quad
  \hstack(\bbA) =
  \begin{bmatrix}
    \bbA_{\bul, \bul, 1} &\cdots& \bbA_{\bul, \bul, K}
  \end{bmatrix},
\end{equation}
where $\bbA_{\bul, \bul, k}$ stands for the $d \times d$ matrix with entries 
$(\bbA_{\bul, \bul, k})_{j, j'} = \bbA_{j, j', k}$.
In view of Equation~\eqref{eq:intensity_hstack}, all the impacts of nodes $j'$ at time scale $k$ on node $j$ is encoded in the $j$-th row of the $d \times Kd$ matrix $\hstack(\bbA)$.
Therefore, a natural assumption is that the matrix $\hstack(\bbA)$ has a low-rank: we assume that there exist latent factors that explain the way nodes impact other nodes through the different scales $k = 1, \ldots, K$.

To induce these prior assumptions on the parameters, we use a
penalization based on a mixture of the $\ell_1$ and trace-norms.
These norms are respectively the tightest convex relaxations for sparsity and
low-rank, see for instance~\cite{Candes04, Candes09}.
They provide state-of-the art results in compressed sensing and
collaborative filtering problems, among many other problems.
These two norms have been previously combined for the estimation of sparse and
low-rank matrices, see for instance~\cite{richard14}
and~\cite{zhou2013learning} in the context of MHP.
Therefore, we consider the following penalization on the parameter $\theta = (\mu, \bbA)$:
\begin{equation}
  \label{eq:penalization}
  \pen(\theta) = \norm{\mu}_{1, \hat w} + \norm{\bbA}_{1, \hat \bbW} +
  \hat \tau \norm{\hstack(\bbA)}_*,
\end{equation}
where each terms are entry-wise weighted $\ell_1$ and trace-norm penalizations given by
\begin{equation*}
  \norm{\mu}_{1, \hat w} = \sum_{j=1}^d \hat w_j |\mu_j|, \quad
  \norm{\bbA}_{1, \hat \bbW} = \sum_{1 \leq j, j' \leq d, 1 \leq k \leq K}
  \hat \bbW_{j, j', k} |\bbA_{j, j', k}|, \quad \norm{\bs A}_* = \sum_{j=1}^d
  \sigma_j(\bs A),
\end{equation*}
where the $\sigma_1(\bs A) \geq \cdots \geq \sigma_d(\bs A)$ are the
singular values of a matrix $\bs A$ (we take $\bs A = \hstack(\bbA)$ in the 
penalization).
The weights $\hat w$, $\hat \bbW$, and coefficients $\hat \tau$ are data-driven tuning 
parameters described below.
The choice of these weights comes from a sharp analysis of the
noise terms and lead to a data-driven scaling of the variability of
information available for each nodes.

From now on, we fix some confidence level $x > 0$, which corresponds to the probability that
the oracle inequality from Theorem~\ref{thm:fast-oracle} holds (see Section~\ref{sec:oracle-inequalities} below).
This can be safely chosen as $x = \log T$ for instance, as described in our numerical experiments (see Section~\ref{sec:numerical_experiment} below).

\paragraph{Weight $\hat \tau$ for the trace-norm penalization of $\hstack(\bbA)$.} % (fold)

This weight comes from Corollary~\ref{cor:deviation-particular-case} 
(see Section~\ref{sec:proof_of_oracles}).
Let us introduce the $d \times Kd$ matrix $\bH(t) = \hstack(\bbH(t))$ where $\bbH(t)$ is 
the $d \times d \times K$ tensor defined by~\eqref{eq:def_bbH} and $\hstack$ is the 
horizontally stacking operator defined by~\eqref{eq:hstack_definition}.
Let us also recall that $\norm{\cdot}_{2}$ is the $\ell_2$-norm, and define 
$\norm{\bH}_{\infty, 2} = \max_{1 \leq j \leq d} \norm{\bH_{j, \bul}}_2$ where 
$\bH_{j, \bul}$ stands for the $j$-th row of $\bH$.
We define
\begin{equation}
  \label{eq:hat_w_star_def}
  \begin{split}
    \hat \tau &= 4 \sqrt{\frac{\lambda_{\max}(\hbV(T) / T) (x + \log(2d) + 
    \ell_\tau(T))}{T}} \\
    & \quad \quad + 28.78 \frac{x + \log(2d) + \ell_\tau(T)) (1 + \sup_{0 \leq t \leq T}
    \norm{\bH(t)}_{\infty, 2})}{T}    
  \end{split}
\end{equation}
where 
\begin{equation*}
  \lmax(\widehat \bV(T)) = \lmax\Big(\int_0^T \bH^\top(s) \bH(s) \diag(d N(s))\Big)
  \; \bigvee \; \max_{j=1, \ldots, d} \int_0^T \norm{\bH_{j, \bul}(t)}_2^2 d N_j(s),
\end{equation*}
and where
\begin{equation*}
  % \label{eq:hat_ell_x_t_def}
  \ell_\tau(T) = 2 \log \log \Big( \frac{4 \lambda_{\max}(\hbV(T))}{x}
  \vee 2 \Big) + 2 \log \log \Big( 4\sup_{0 \leq t \leq T}
  \norm{\bH(t)}_{\infty, 2} \vee 2 \Big),
\end{equation*}
where we used the notation $a \vee b = \max(a, b)$ for $a, b \in \R$.

\paragraph{Weights $\hat w_{j}$ for $\ell_1$-penalization of $\mu$.}

These weights are given by
\begin{equation}
  \label{eq:hat_w_def} 
  \hat w_j = 6 \sqrt{\frac{(N_j(T) / T) (x + \log d + 
  \ell_j(T))}{T}} + 86.34 \frac{x + \log d + \ell_j(T)}{T}
\end{equation}
with $\ell_j(T) = 2 \log \log (\frac{4 N_j(T)}{x} \vee 2 ) ) + 2 \log \log 4$.
The weighting of each coordinate $j$ in the penalization of $\mu$ is
natural: it is roughly proportional to the square-root of $N_j(T)
/ T$, which is the average intensity of events on coordinate $j$. The
term $\ell_{j}(T)$ is a technical term, that can be neglected
in practice, see Section~\ref{sec:numerical_experiment}.

\paragraph{Weights $\hat \bbW_{j, j' k}$ for $\ell_1$-penalization of $\bbA$.} % (fold)

Recall that the tensor $\bbH$ is given by~\eqref{eq:def_bbH}.
The weights $\hat \bbW_{j, j' k}$ are given by
\begin{equation}
  \label{eq:hat_W_def}
  \begin{split}
    \hat \bbW_{j,j', k} &= 4 \sqrt{\frac{\frac 1T \int_0^T \bbH_{j, j', k}(t)^2 d N_j(t) 
  (x + \log(K d^2) + \mathbb L_{j, j', k}(T))}{T}} \\
  & \quad + 28.78 \frac{(x + \log(K d^2) + \mathbb L_{j,j', k}(T))(1 + 
  \sup_{0 \leq t \leq T} |\bbH_{j, j', k}(t)|)}{T}
  \end{split}
\end{equation}
where $\mathbb L_{j,j', k}(T) = 2 \log \log \big( \frac{4 \int_0^T \bbH_{j, j', k}(t)^2 d N_j(t)}{x} \vee 2 \big) + 2 \log \log (4 \sup_{0 \leq t \leq T} |\bbH_{j, j', k}(t)| \vee 2)$.
Once again, this is natural: the variance term $\int_0^T \bbH_{j, j', k}(t)^2 d N_j(t)$
is, roughly, an estimation of the variance of the self-excitements between coordinates $j$ and $j'$ at time scale $k$.
The term $\mathbb L_{j, j', k}(T)$ is a technical term that can be neglected in practice.

These weights are actually quite natural: the terms $\lambda_{\max}(\hbV(T))$ and
$\int_0^T \bbH_{j, j', k}(t)^2 d N_j(t)$ correspond to estimations of the noise variance, 
that are the $L^2$ terms appearing in the empirical Bernstein's inequalities given in
Section~\ref{sec:concentration-inequalities}. 
The terms $\sup_{0 \leq t \leq T} \norm{\bH(t)}_{\infty, 2}$ and 
$\sup_{0 \leq t \leq T} |\bbH_{j, j', k}(t)| $ correspond to the
$L^\infty$ terms from these Bernstein's inequalities.
Once again, these data-driven weights lead to a sharp tuning of the penalizations, as illustrated numerically in Section~\ref{sec:numerical_experiment} below.

\section{A sharp oracle inequality}
\label{sec:oracle-inequalities}

Recall that the inner product $\inr{\lambda_1, \lambda_2}_T$ is given
by~\eqref{eq:inner-product} and recall that $\norm{\cdot}_T$ stands
for the corresponding norm. 
Theorem~\ref{thm:fast-oracle} below is
a sharp oracle inequality on the prediction error measured by
$\norm{\lambda_{\hat \theta} - \lambda}_T^2$. For the proof of oracle
inequalities with a fast rate, one needs a restricted eigenvalue
condition on the Gram matrix of the problem~\citep{MR2533469,
  koltchinskii2011oracle}. One of the weakest assumptions considered
in literature is the Restricted Eigenvalue (RE) condition. 
In our setting, a natural RE assumption is given in Definition~\ref{ass:RE}
below.
First, we need to introduce some simple notations and definitions.

\paragraph{Some notations and definitions.}

If $a, b$ (resp. $\bA, \bB$ and $\bb A, \bb B$) are vectors (resp. matrices and tensors) 
of the same size, we always denote by $\inr{a, b}$ (resp. $\inr{\bA, \bB}$ and $\inr{\bb A, \bb B}$) their inner products.
For matrices this can be written as $\inr{\bA, \bB} = \sum_{i, j} \bA_{i, j} \bB_{i, j} 
= \tr(\bA^\top \bB)$, where $\tr$ stands for the trace, while for (say, three dimensional) tensors we write similarly $\inr{\bb A, \bb B} = \sum_{i, j, k} \bb A_{i, j, k} \bb B_{i, j, k}$.
We define the Euclidean norm (Frobenius) for tensors and matrices simply as 
$\norm{\bA}_F = \sqrt{\inr{\bA, \bA}}$ and $\norm{\bbA}_F = \sqrt{\inr{\bbA, \bbA}}$.
If $\bW$ (resp. $\bbW$) is a matrix (resp. tensor) with positive entries, we introduce 
the weighted entrywise $\ell_1$-norm
given by $\norm{\bA}_{1, \bW} = \inr{\bW, |\bA|}$, 
(resp.  $\norm{\bbA}_{1, \bbW} = \inr{\bbW, |\bbA|}$) where $|\bA|$ (resp. $|\bbA|$)
contains the absolute values of the entries of $\bA$ (resp. $\bbA$). 
If $A$ is a vector, matrix or tensor then $\norm{A}_0$ is the number of non-zero entries 
of $A$, while $\supp(A)$ stands for the support of $A$ (indices of non-zero entries)
For another vector, matrix or tensor $A'$ with the same shape, the
notation $[A']_{\supp(A)}$ stands for the vector, matrix or tensor with the same 
coordinates as $A'$ where we put $0$ at indices outside of $\supp(A)$.
We also use the notation $u \vee v = \max(u, v)$ for $a, b \in \R$.

If $\bA = \bU \bSigma \bV^\top$ is the SVD of a $m \times n$ matrix $\bA$, with the 
columns $u_j$ of $\bU$ and $v_k$ of $\bV$ being, respectively, the orthonormal left and right
singular vectors of $\bA$, the projection matrix onto the space
spanned by the columns (resp. rows) of $\bA$ is given by $\bP_{\bU} =
\bU \bU^\top$ (resp. $\bP_{\bV} = \bV \bV^\top$). The operator
$\cP_{\bA} : \R^{m \times n} \rightarrow \R^{m \times n}$ given by
$\cP_{\bA}(\bB) = \bP_{\bU} \bB + \bB \bP_{\bV} - \bP_{\bU} \bB
\bP_{\bV}$ is the projector onto the linear space spanned by the
matrices $u_j x^\top$ and $y v_k^\top$ for all $1 \leq j, k \leq \rank(\bA)$ and
$x \in \R^n$, $y \in \R^{m}$. 
The projector onto the orthogonal space is given by
$\cP_{\bA}^\perp(\bB) = (\bI - \bP_{\bU}) \bB (\bI - \bP_{\bV})$.

\begin{definition}
  \label{ass:RE}
  Fix $\theta = (\mu, \bbA)$ where $\mu \in \R^d$ and $\bbA \in \R_+^{d
    \times d \times K}$ and define $\bA = \hstack(\bbA)$. 
  We define the constant $\kappa(\theta) \in (0, +\infty]$ such that, for
  any $\theta' = (\mu', \bbA')$ and $\bA' = \hstack(\bbA')$ satisfying
  \begin{align*}
  \frac 13 &\norm{(\mu')_{\supp(\mu)^\perp}}_{1, \hat w} 
  + \frac 12 \norm{(\bbA')_{\supp({\bbA})^\perp}}_{1, \hat \bbW}
  + \frac 12 \hat \tau \norm{\cP_{\bA}^\perp(\bA')}_* \\
   &\leq \frac 53 \norm{(\mu')_{\supp(\mu)}}_{1, \hat w}
   + \frac 32 \norm{(\bbA')_{\supp({\bbA})}}_{1, \hat \bbW}
   + \frac 32 \hat \tau \norm{\cP_{\bA}(\bA')}_*,
\end{align*}
  % \begin{equation*}
  %   \norm{(\mu')_{\supp(\mu)^\perp}}_{1, \hat w} \leq 5
  %   \norm{(\mu')_{\supp(\mu)}}_{1, \hat w}
  % \end{equation*}
  % and
  % \begin{align*}
  %   \norm{(\bbA')_{\supp(\bbA)^\perp}}_{1, \hat \bbW} + \hat \tau
  %   \norm{\cP_{\bA}^\perp(\bA')}_* \leq 3
  %   \norm{(\bbA')_{\supp(\bbA)}}_{1, \hat \bbW} + 3 \hat \tau
  %   \norm{\cP_{\bA}(\bA')}_*,
  % \end{align*}
  we have
  \begin{equation*}
    % \label{eq:using_RE}
    \norm{(\mu')_{\supp(\mu)}}_{2} \vee \norm{(\bbA')_{\supp(\bbA)}}_{F}
    \vee \norm{\cP_\bA(\bA')}_{F} \leq \kappa(\theta)
    \norm{\lambda_{\theta'}}_T.
  \end{equation*}
\end{definition}
The constant $1 / \kappa(\theta)$ is a restricted eigenvalue depending
on the ``support'' of $\theta$, which is naturally associated with the
problem considered here. Roughly, it requires that for any parameter
$\theta'$ that has a support close to the one of $\theta$ (measured by
domination of the $\ell_1$ norms outside the support of $\theta$ by
the $\ell_1$ norm inside it), we have that the $L^2$ norm of the
intensity given by $\norm{\lambda_{\theta'}}_T$ can be compared with
the $L^2$ norm of $\theta'$ in the support of $\theta$.
Note that for a given $\theta$, we simply allow $\kappa(\theta) = +\infty$, so the restricted eigenvalue is zero, whenever the inequality is not met (which makes in such as case the statement of Theorem~\ref{thm:fast-oracle} trivial).

\begin{theorem}
  \label{thm:fast-oracle}
  Fix $x > 0$, and let $\hat \theta$ be given
  by~\eqref{eq:hat_theta_def_L2_risk} and~\eqref{eq:penalization} with tuning parameters given
  by~\eqref{eq:hat_w_star_def},~\eqref{eq:hat_w_def} and~\eqref{eq:hat_W_def}. 
  Then, the inequality
\begin{equation}
  \begin{split}
    \norm{\lambda_{\hat \theta} - \lambda}_T^2 \leq \inf_{\theta = (\mu, \bbA)}
    \bigg\{ \norm{\lambda_{\theta} - \lambda}_T^2 & + 1.25 \kappa(\theta)^2
    \Big(\norm{(\hat w)_{\supp(\mu)}}_2^2 \\
    & + \norm{(\hat \bbW)_{\supp(\bbA)}}_F^2 + \hat \tau^2 \rank(\hstack(\bbA)) 
    \Big) \bigg \}
  \end{split}
\end{equation}
  holds with a probability larger than $1 - 70.35 e^{-x}$.
\end{theorem}

The proof of Theorem~\ref{thm:fast-oracle} is given in 
Section~\ref{sec:proof_of_oracles} below.
Note that no assumption is required on the ground truth intensity
$\lambda$ of the multivariate counting process $N$ in Theorem~\ref{thm:fast-oracle}.
Moreover, if one forgets in Section~\ref{sec:procedure} about the negligible terms 
$\ell_\tau(T), \ell_j(T)$ and $\bb L_{j, j', k}(T)$ and if one keeps only the dominating
$L^2$ terms in $O(1 / T)$ (while $L^\infty$ terms are $O(1 / T^2)$ in the 
large $T$ regime), we obtain upper bounds, up to numerical constants (denoted $\lesssim$), for 
the terms involved in Theorem~\ref{sec:oracle-inequalities}:
\begin{equation*}
  \norm{(\hat w)_{\supp(\mu)}}_2^2 \lesssim \norm{\mu}_0 \max_{j \in
    \supp(\mu)} \frac{\frac 1T N_j(T) (x + \log d)}{T},
\end{equation*}
where $\norm{\mu}_0$ stands for the sparsity of $\mu$,
\begin{align*}
  \norm{(\hat \bbW)_{\supp(\bbA)}}_F^2 &\lesssim \norm{\bbA}_0 \max_{(j, j', k) \in
    \supp(\bbA)} \frac{ \frac 1T \int_0^T \bb H_{j, j', k}(t)^2 d N_j(t) (x + \log(K d^2))}{T},
\end{align*}
where $\norm{\bbA}_0$ stands for the sparsity of $\bbA$, and finally
\begin{equation*}
    \hat \tau^2 \lesssim \rank (\hstack(\bb A)) \frac{\frac 1T \lambda_{\max}(\hbV(T))
     (x + \log(2d))}{T}.
\end{equation*}
Hence, Theorem~\ref{thm:fast-oracle} proves that $\hat \theta$ achieves an
optimal trade-off between approximation and complexity, where the
complexity is, roughly, measured by
\begin{align*}
  \frac{\norm{\mu}_0 (x + \log d)}{T} & \max_{j} \frac{N_j(T)}{T} \; + \;
  \frac{\norm{\bbA}_0 (x + \log (K d^2))}{T} \max_{j, j', k} 
  \frac 1T \int_0^T \bb H_{j, j', k}(t)^2 d N_j(t) \\
  & \quad + \frac{\rank(\hstack(\bbA)) (x + \log(2d))}{T} \frac 1T \lmax(\widehat \bV(T)).
\end{align*}
Note that typically $K \leq d$ so that $\log(K d^2) \leq 3 \log d$, which means 
that $\log (K d^2)$ scales as $\log d$. 
The complexity term depends on both the sparsity of $\bbA$ and the rank of
$\hstack(\bbA)$. 
The rate of convergence has the ``expected'' shape $(\log d) / T$, recalling that 
$T$ is the length of the observation interval of
the process, and these terms are balanced by the empirical variance
terms coming out of the new concentration results given in 
Section~\ref{sec:concentration-inequalities} above.

\section{Numerical experiments}
\label{sec:numerical_experiment}

In this Section we conduct experiments on synthetic datasets to evaluate the performance of our
method, based on the proposed data-driven weighting of the penalizations, compared to unweighted
penalizations~\citep{zhou2013learning}.
Throughout this Section, we consider the most widely used sum of exponentials kernel, defined in Equation~\eqref{eq:sum_of_exponentials_kernel}.

\subsection{Simulation setting}

\begin{figure}[htbp]
  \centering
  \includegraphics[width=0.9\textwidth]{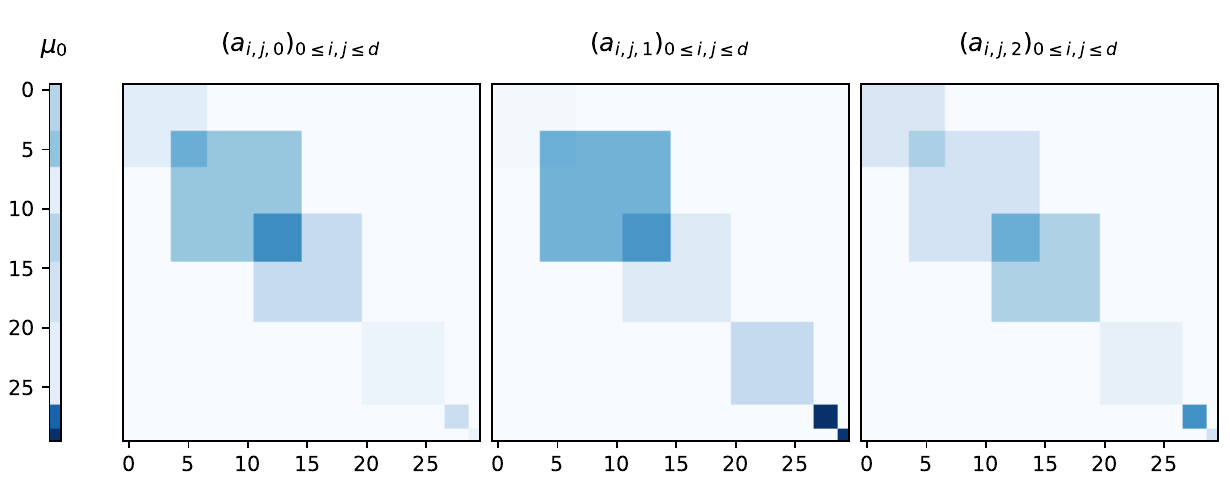}
  \includegraphics[width=0.9\textwidth]{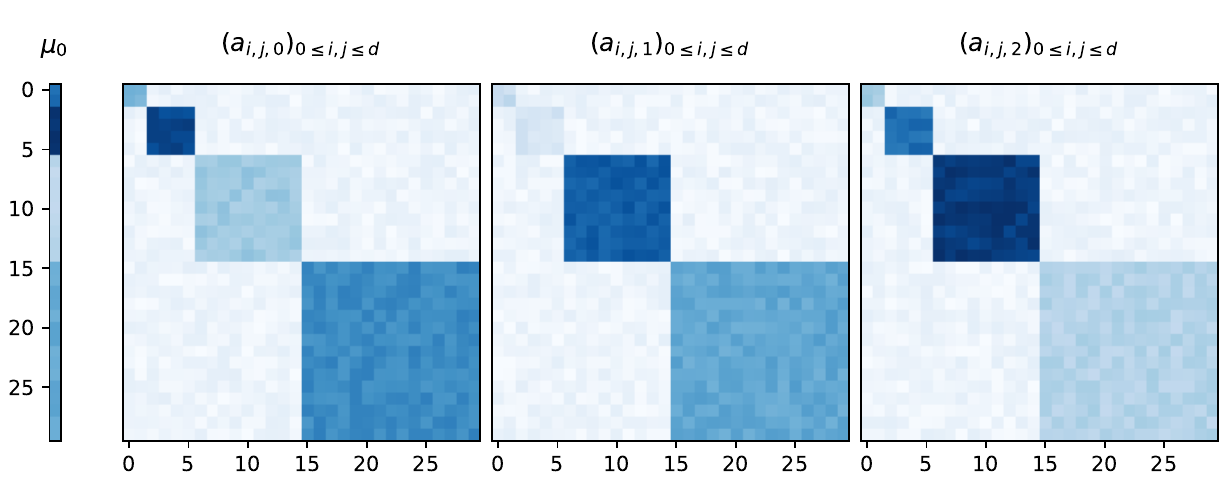}
  \includegraphics[width=0.9\textwidth]{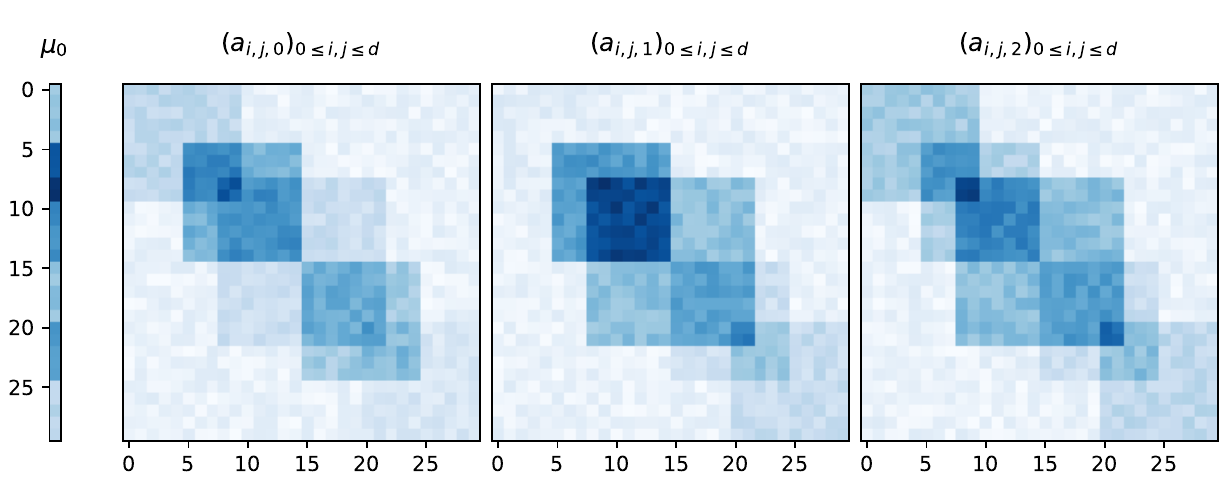}
  \caption{Ground truth vector $\mu$ and tensor $\bbA$ in dimension 30. Each row corresponds to a different example used in our experiments.}
  \label{fig:ground_truth}
\end{figure}

We generate Hawkes processes using Ogata's thinning algorithm~\citep{ogata1981lewis} with $d = 30$ nodes.
Baseline intensities $\mu_j$ are constant on blocks, we use $K=3$ basis kernels
$h_{j, j', k}(t) = \alpha_k e^{-\alpha_k t}$ with $\alpha_1 = 0.5$, $\alpha_1 = 2$ 
and $\alpha_3 = 5$.
We consider three examples for the slices $\bbA_{\bul, \bul, 1}$, $\bbA_{\bul, \bul, 2}$ and $\bbA_{\bul, \bul, 3}$ of the adjacency tensor $\bbA$, including settings with overlapping boxes, and noisy entries over the block structure, as illustrated in Figure~\ref{fig:ground_truth}.
These blocks correspond to the overlapping communities reacting at different time scales.
The tensor $\bbA$ is rescaled so that the operator norm of the matrix 
$\sum_{k=1}^3 \bbA_{\bul, \bul, k}$ is equal to $0.8$, guaranteeing to obtain a 
stationary process.
For each simulated data, we increase the length of the time interval $T = 5000, 7000, 10000, 15000, 20000$, and fit each time the procedures.
An overall averaging of the results is computed on 100 separate simulations.

\subsection{Procedures and metrics} % (fold)

We consider a procedure based on the minimization of the least-squares functional~\eqref{eq:least-squares}.
This objective is convex, with a goodness-of-fit term which is gradient-Lipschitz: we use first-order
optimization algorithms, based on proximal gradient descent.
Namely, we use Fista~\citep{fista} for problems with a single penalization on $\bbA$
($\ell_1$-norm or trace norm penalization of $\hstack(\bbA)$) and GFB (generalized forward backward, see~\cite{pino1999generalized}) for mixed $\ell_1$ penalization of $\bbA$ and trace-norm penalization of $\hstack(\bbA)$.
For both procedures we choose a fixed gradient step equal to $1/L$ where $L$ is the Lipschitz
constant of the loss, namely the largest singular value of the Hessian (which is constant for this least-squares functional).
We limit our algorithms to $25,000$ iterations and stop when the objective relative decrease is 
less than $10^{-10}$ for Fista and $10^{-7}$ for GFB.
We only penalize $\bbA$ and consider the following procedures:
\begin{itemize}
% \item NoPen : direct minimization of the least-squares, with no penalization;
\item L1: non-weighted L1 penalization;
\item wL1: weighted L1 penalization;
\item Nuclear: non-weighted trace-norm penalization;
\item L1Nuclear: non-weighted L1 penalization and trace-norm penalization;
\item wL1Nuclear: weighted L1 penalization and trace-norm penalization.
\end{itemize}
Note that L1Nuclear is the same as the procedure considered in~\cite{zhou2013learning},
however, we use a different optimization algorithm, based on an
proximal gradient descent (a first-order method, which is typically faster than an algorithm based on ADMM, as proposed in~\cite{zhou2013learning}).
The data-driven weights used in our procedures are the ones derived from our
analysis, see~\eqref{eq:hat_w_star_def} and~\eqref{eq:hat_W_def}, where we simply put
$x = \log T$.
For each metric, we tune the constant in front the $\ell_1$ penalization, and the constant in front of the trace-norm penalization in order to obtain the best possible metrics for each procedure, on average over all separate
simulations.
Namely, there is no test set, we simply display the best metrics obtained by each procedure
for a fair comparison.
All experiments are done using our \texttt{tick} library for \texttt{Python3}, see~\cite{tick}, its GitHub page is \url{https://github.com/X-DataInitiative/tick} and documentation is available here~\url{https://x-datainitiative.github.io/tick/}.
The following metrics are considered in order to assess the procedures.
\begin{description}
\item[Estimation error:] the relative $\ell_2$ estimation error of $\bbA$, given by
  $\norm{\hat \bbA - \bbA}_2^2 / \norm{\bbA}_2^2$
\item [AUC:] we compute the AUC (area under the ROC curve) between the binarized ground truth matrix
  $\bbA$ and the solution $\hat \bbA$ with entries scaled in $[0, 1]$.
  This allows us to quantify the ability of the procedure to detect the support of the connectivity
  structure between nodes.
\item [Kendall:] we compute Kendall's tau-b between all entries of the ground truth matrix $\bbA$ and
  the solution $\hat \bbA$.
  This correlation coefficient takes value between $-1$ and $1$ and compare the number of concordant and
  discordant pairs.
  This allows us to quantify the ability of the procedure to rank correctly the intensity of the
  connectivity between nodes.
\end{description}

\subsection{Results} % (fold)

\begin{figure}[htbp]
  \centering
  \includegraphics[height=0.8\textheight]{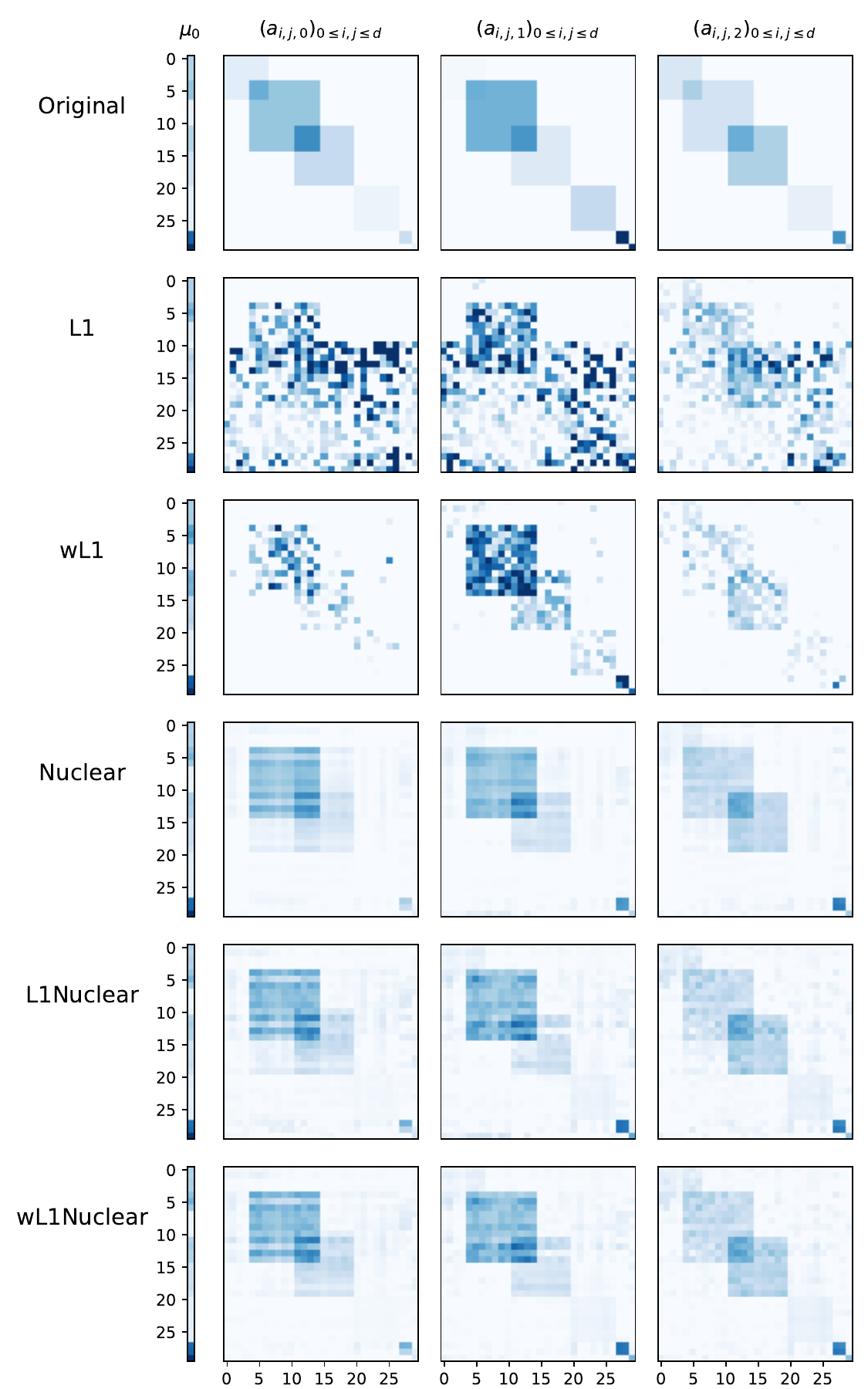}
  \caption{Ground truth tensor $\bbA$ and recovered tensors using all procedures.
    We observe that wL1 and wL1Nuclear leads to a much better support recovery, since we observe less false positives outside of the node communities.}
  \label{fig:matrices}
\end{figure}

\begin{figure}[htbp]
  \raggedleft
  \includegraphics[width=0.75\textwidth]{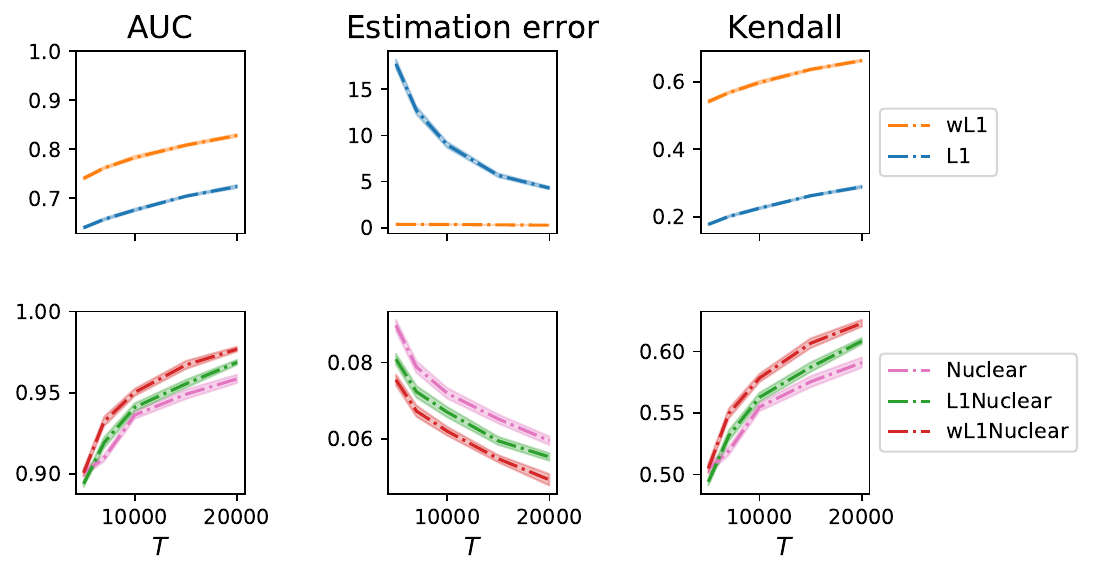}
  \includegraphics[width=0.75\textwidth]{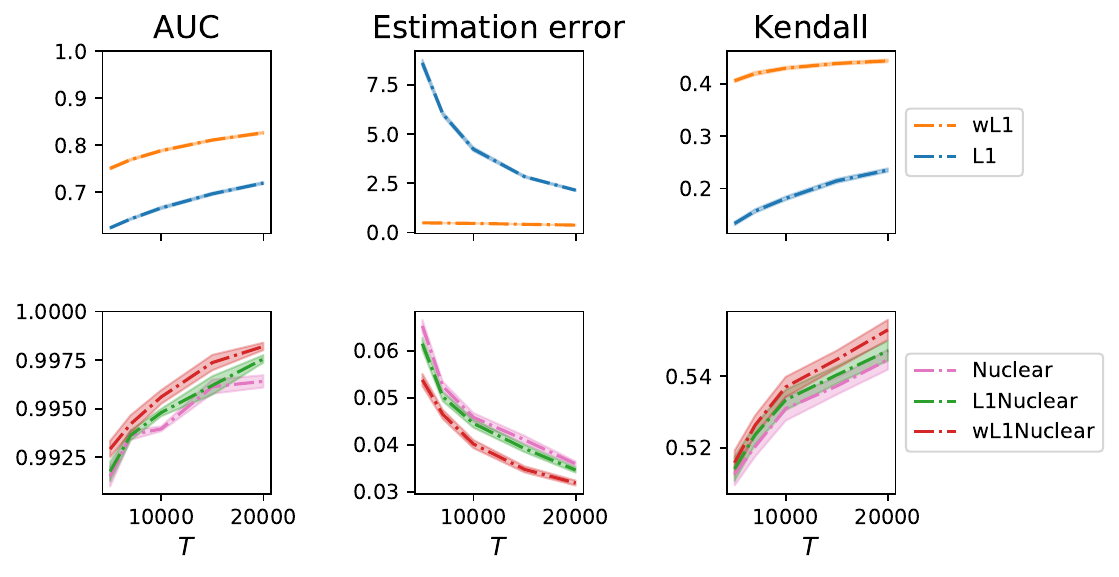}
  \includegraphics[width=0.75\textwidth]{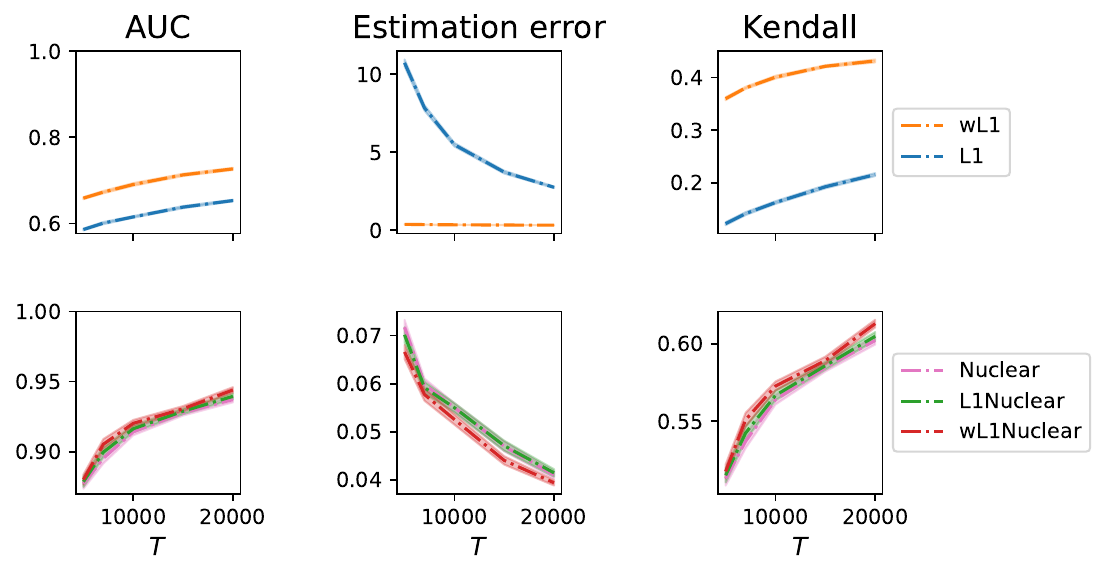}
  \caption{Average metrics achieved by all procedures on the three considered examples of $\bbA$ (in the same order as the display from Figure~\ref{fig:ground_truth}), and 95\% confidence bands, with increasing observation length $T$ over repeated simulations.
  Weighted penalizations systematically lead to improvements over L1, Nuclear and L1 + Nuclear penalization.}
  \label{fig:metrics}
\end{figure}

In Figure~\ref{fig:matrices} we observe, on an instance of the problem, the strong improvements of wL1 and wL1Nuclear over L1, Nuclear and L1Nuclear respectively.
We observe in particular that a sharp tuning of the penalizations, using data-driven weights, leads to a much smaller number of false positives outside the node communities (better viewed on a computer).
In Figure~\ref{fig:metrics}, we compare all the procedures in terms of estimation error, AUC and
Kendall coefficient and confirm the fact that weighted penalizations systematically lead to an
improvement, both over unweighted L1, Nuclear and L1Nuclear.

\subsection{A comparison of the least-squares and likelihood functionals} % (fold)

This paper considers, mostly for theoretical reasons, least-squares as a goodness-of-fit for the Hawkes process.
However, estimation in this model is usually achieved by minimizing the goodness-of-fit given by the negative log-likelihood.
In what follows, we provide some numerical insights in order to compare objectively both approaches.

\begin{figure}[htbp]
  \centering
  \includegraphics[width=0.9\textwidth]{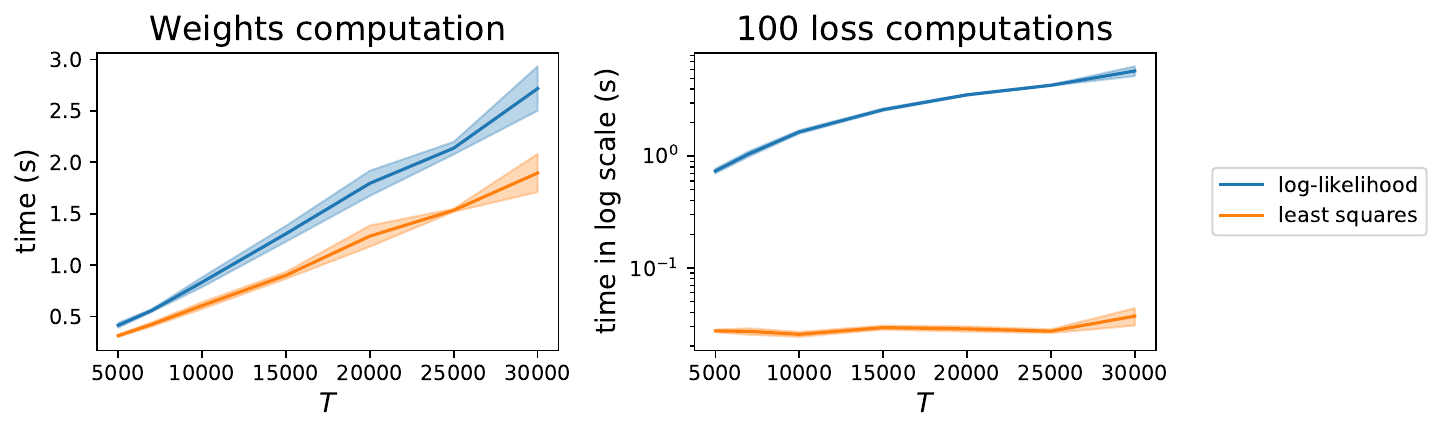}%
  \caption{Average time needed for weights (left) and value computation (right) (and 95\% confidence bands) for least squares and log-likelihood with precomputations, over repeated simulations. 
  We observe that value computations are order of magnitude faster for least-squares ($y$-scale is logarithmic on the right hand side) and constant with an increasing observation length, while it is strongly increasing for the log-likelihood.}
  \label{fig:contrast_vs_llh_complexity}
\end{figure}

First, one can precompute for both functionals some weights in order to accelerate future
gradient and value computations.
In both cases, the precomputations have similar complexities, unless the number of kernels $K$ is large (see Table~\ref{table:contrast_vs_llh_complexity} below).
However, given such precomputations, a remarkable property of the least-squares versus the log likelihood is that value and gradient computation is independent of the total number of observed events (denoted $n$): complexity is $O(K^2 d^3)$ for least-squares, while it is $O(n K d)$ for log likelihood, which means that such computations for least-squares can be orders of magnitude faster whenever $n \gg K d^2$, which is the case in the setting considered in our experiments.
For instance, experiments used to produce Figures~\ref{fig:matrices} and~\ref{fig:metrics} for $T = 20,000$ use about $n \approx 500,000$ events, and $d=30, K=3$.
Note that, however, the least-squares approach considered here does not scale with respect to $d$ because of its $O(d^3)$ complexity, we recommend to use instead the negative log-likelihood whenever $d$ is large (larger than $1000$, say).
The complexity of each operation is described in Table~\ref{table:contrast_vs_llh_complexity} below and a numerical illustration of this complexity is displayed in Figure~\ref{fig:contrast_vs_llh_complexity}, which confirms that computations with least-squares are orders of magnitude faster than with log-likelihood in the considered setting.
We don't provide proofs for these complexities, since it follows straightforward arguments, however details about this can be found in Chapter~2 of~\cite{bompaire_phd}.

\begin{table}[htbp]
\centering{}
\begin{tabular}{ c | c c c c }
               & pre-computation & memory & value & gradient \\
 \hline
 Least squares & $O(n K^2 d)$ & $O(K^2 d^3)$ & $O(K^2 d^3)$ & $O(K^2 d^3)$ \\
 Likelihood    & $O(n K d)$  & $O(n K d)$  & $O(n K d)$  & $O(n K d)$
\end{tabular}
\caption{From left to right:
  Weights precomputation complexity, memory storage, value and gradient complexity for both functionals.
  Note that for least-squares, the complexity of the value and the gradient with
  precomputed weights is independent on the number of events~$n$.}
\label{table:contrast_vs_llh_complexity}
\end{table}
Another important point is related to smoothness properties: the negative log-likelihood does not satisfy the gradient-Lipschitz assumption, while this property is required by most first order optimization algorithms to obtain convergence guarantees and an easy tuning of the step-size used in gradient descent.
Therefore, for the negative log-likelihood, convergence can be very unstable, while on the contrary, least-squares is gradient Lipschitz and is easy to optimize since it is a quadratic function.
Note that in~\cite{bompaire2018dual} is proposed an alternative approach based on duality, in particular for the negative log-likelihood of the Hawkes process.
Herein one can observe the strong instability of standard first order algorithms (such as the one considered here) for the negative log-likelihood.

In Figure~\ref{fig:contrast_vs_llh_loss_convergence} below, we compare the
performances of ISTA and FISTA with linesearch for automatic step-size tuning, both for least-squares and negative log-likelihood.
This figure confirms that the number of iterations required for least-squares
is much smaller than for the negative log-likelihood.
This gap is even stronger if we look at the computation times, since each iteration is computationally faster with least squares, and even more so when the observation length increases.
\begin{figure}[htbp]
  \centering
  \includegraphics[width=0.85\textwidth]{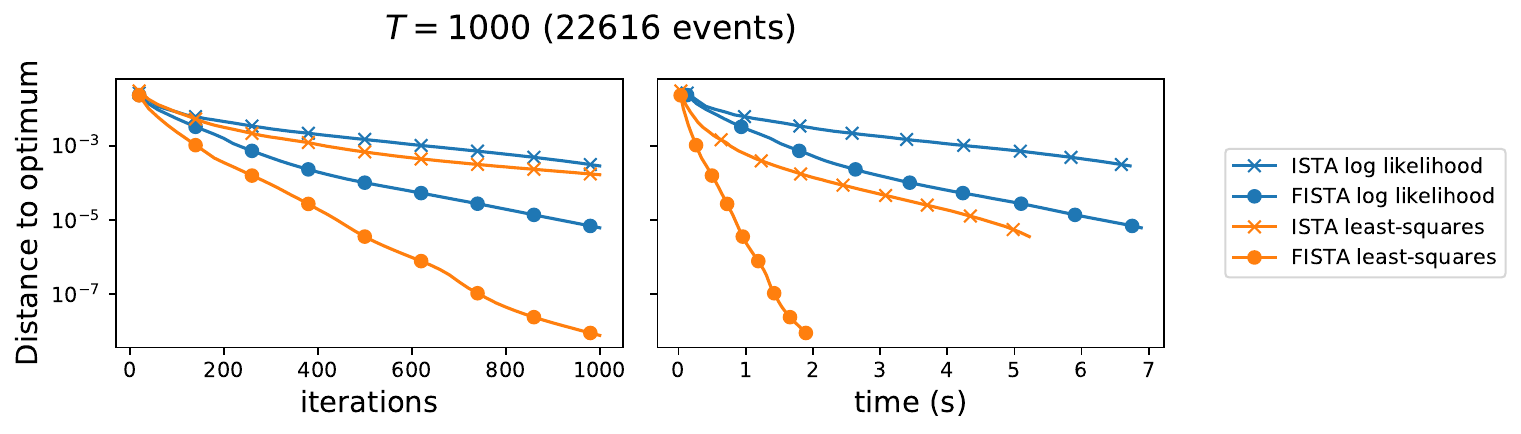} \\
  \vspace{0.3cm}
  \includegraphics[width=0.85\textwidth]{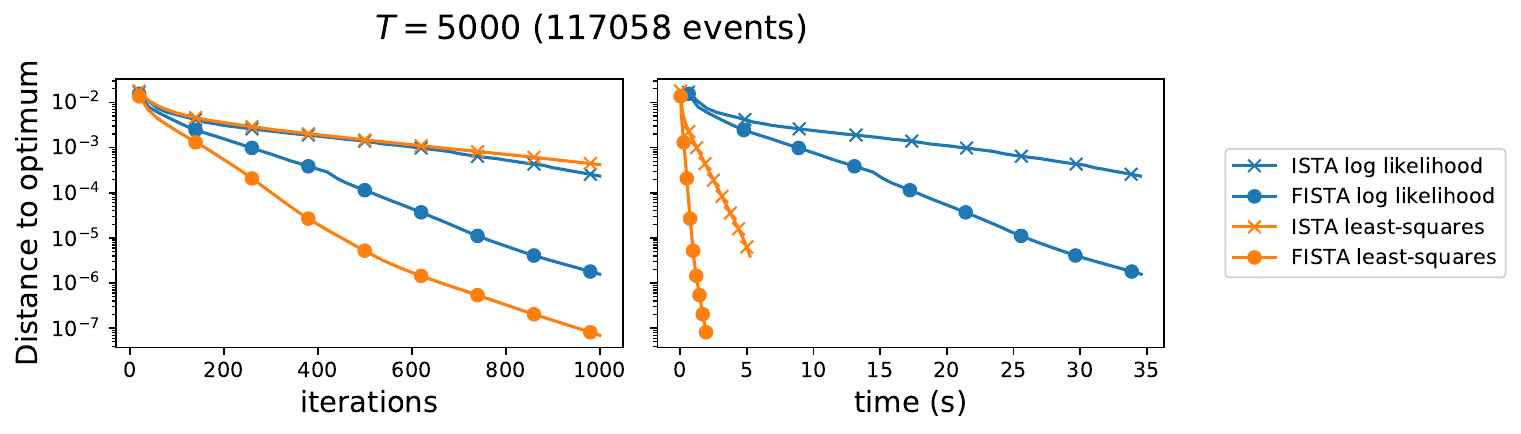}%
  \caption{Convergence speed of least squares and likelihood losses with ISTA
  and FISTA optimization algorithms on two simulations of a Hawkes process with parameters from
  Figure~\ref{fig:ground_truth} with observation length $T=1000$ (top) and $T=5000$ (bottom).
  Once again, we observe that the computations are much faster with least-squares, in particular with a large observation length.}
  \label{fig:contrast_vs_llh_loss_convergence}
\end{figure}

In this Section, we compared least-squares and log-likelihood for the Hawkes process through
a computational perspective only, and concluded that least-squares is typically order of magnitude faster.
Now, let us compare the statistical performances of both approaches on the same simulation setting as before, with $T=20,000$, using the metrics defined above, namely Estimation Error, AUC and Kendall.
We simply use for this L1 penalization on $\bbA$, with a strength parameter
tuned for each metric and for each goodness-of-fit.
\begin{figure}[htbp]
  \centering
  \includegraphics[width=\textwidth]{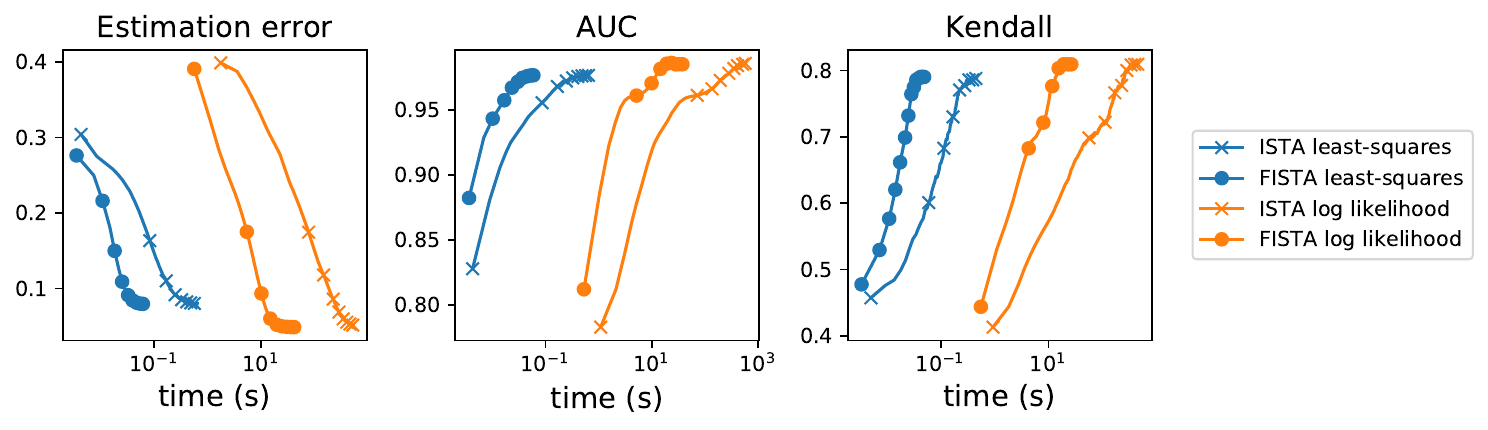}%
  \caption{Metrics achieved by least squares and log-likelihood estimators
  after precomputations. We observe that log-likelihood achieves a slightly better AUC and Estimation Error, but at a stronger computational cost ($x$-axis are on a logarithmic scale).}
  \label{fig:contrast_vs_llh_metrics}
\end{figure}

In Figure~\ref{fig:contrast_vs_llh_metrics}, we observe that both functionals roughly achieve the same performance measured by the Kendall coefficient, but that the negative log-likelihood achieves a slightly better AUC and estimation error than least-squares, at a stronger computational cost.
The slightly better statistical performance of maximum likelihood is not surprising, since vanilla maximum likelihood is known to be statistically efficient asymptotically for Hawkes processes, see~\cite{ogata1978asymptotic}, while up to our knowledge, vanilla least-squares estimator is not.
This leads to the conclusion that least squares are a very good alternative to maximum
likelihood when dealing with a large number of events: statistical accuracy is only slightly deteriorated, but the computational cost is order of magnitudes smaller, and convergence is much more stable.

\begin{figure}[htbp]
  \centering
  \includegraphics[width=\textwidth]{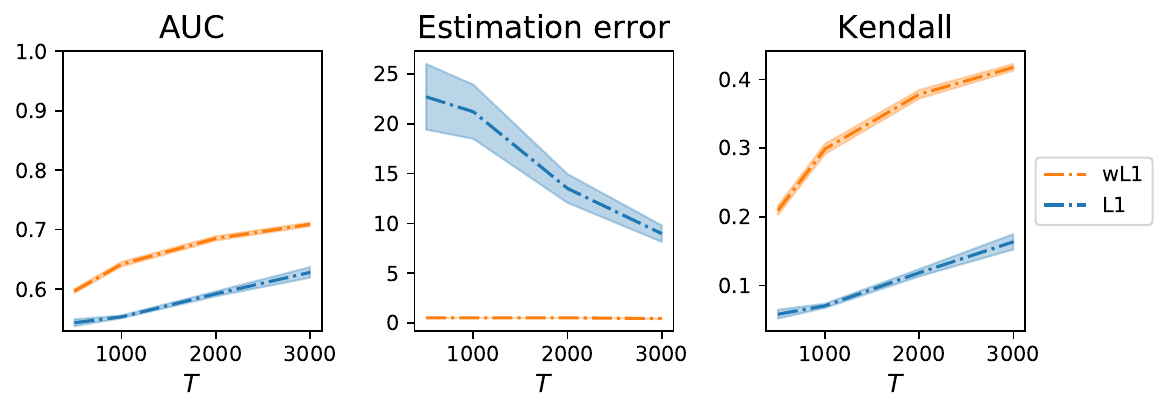}%
  \caption{Performances of $\ell_1$ versus weighted-$\ell_1$ for estimators based on the negative log-likelihood functional, where the data-driven weights used in the $\ell_1$ penalization are the ones derived for the least-squares functional. We observe that these weights allow to improve significantly the performances of $\ell_1$-penalized estimators based on the log-likelihood functional, for all the considered metrics. This is expected, since both functionals are actually close to each other.}
  \label{fig:weights-llh}
\end{figure}

In Figure~\ref{fig:weights-llh}, we observe the performances achieved by $\ell_1$ versus weighted-$\ell_1$ for the estimators based on the log-likelihood functional. 
The point here is that we use the weights $\hat \bbW$ from Equation~\eqref{eq:hat_W_def} that are derived for the least-squares functional.
We observe that, however, these data-driven weights allow to strongly improve over the vanilla $\ell_1$-penalization for the negative log-likelihood estimator as well.
This behavior is actually expected, since both functionals are actually close to each other, and the least-squares functional can even be understood as an approximation of the negative log-likelihood one, see~\cite{bacry2016mean}.

\subsection{Sensitivity to the penalization level and weights}

\begin{figure}[htbp]
  \center
  \includegraphics[width=1.1\textwidth]{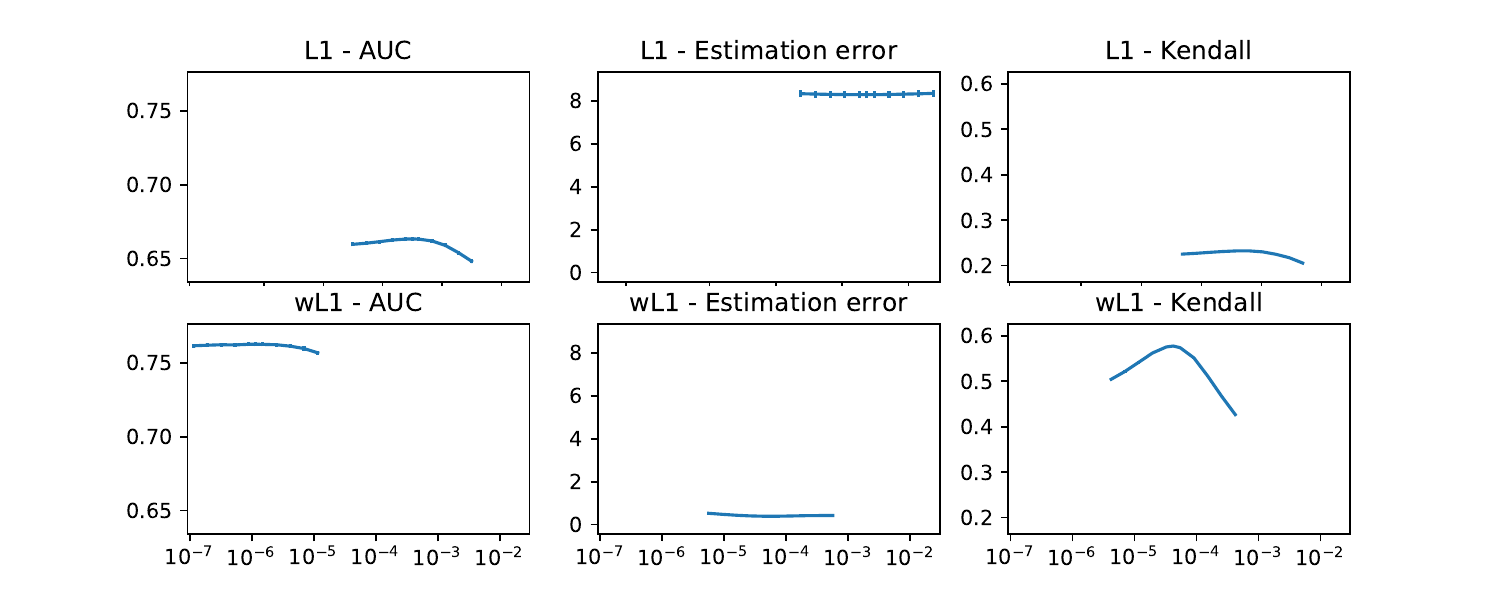}
  \caption{Sensitivity of the metrics (top: AUC, middle: Estimation error, bottom: Kendall) with respect to the penalization level both for unweighted (left-hand side) and weighted (right-hand side) $\ell_1$ penalizations. Weighted $\ell_1$-penalization is more sensitive to its unweighted counterpart, but leads to much better performances even if not perfectly tuned.}
  \label{fig:sensitivity}
\end{figure}

In Figure~\ref{fig:sensitivity}, we display the values of the metrics as a function of the penalization level used, both for unweighted and weighted $\ell_1$ penalization.
We observe that the weighted $\ell_1$-penalization is more sensitive to its unweighted counterpart, but leads anyway to much better performances even if the penalization level is not perfectly tuned.

\begin{figure}[htbp]
  \centering
  \includegraphics[width=0.9\textwidth]{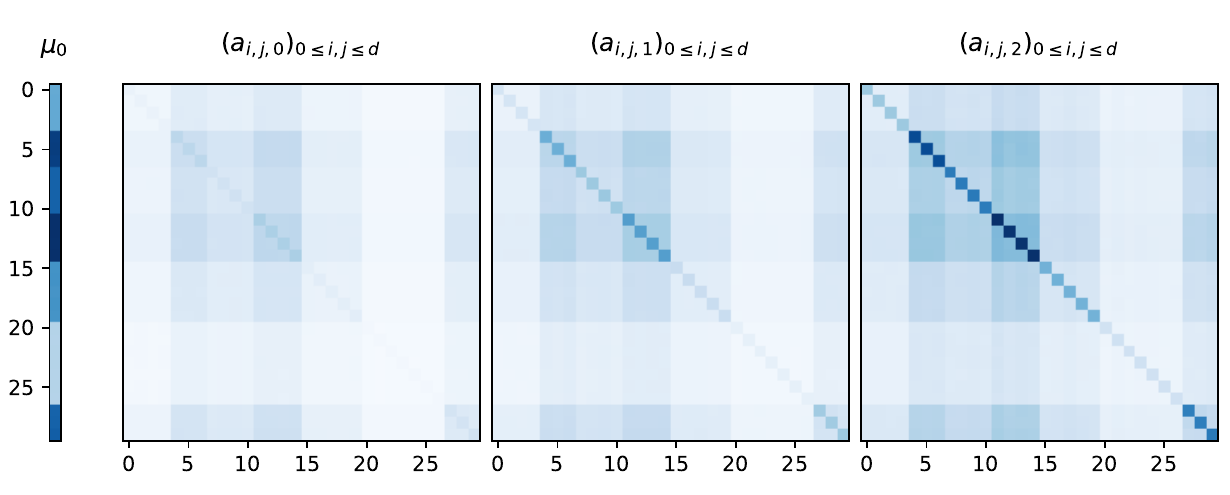}
  \caption{Visualization of the weights used in the weighted-$\ell_1$ penalization for a single simulation from the first setting (corresponding to tensor $\bbA$ displayed in the first row of Figure~\ref{fig:ground_truth}). This corresponds to the weights from Equation~\eqref{eq:hat_W_def}, namely $\hat \bbW_{\bullet, \bullet, 1}$ (left), $\hat \bbW_{\bullet, \bullet, 2}$ (middle) and $\hat \bbW_{\bullet, \bullet, 3}$ (right).}
  \label{fig:penalization_weights}
\end{figure}

In Figure~\ref{fig:penalization_weights} we display the weights $\hat \bbW$ from Equation~\eqref{eq:hat_W_def} used in the weighted-$\ell_1$ penalization for a single simulation from the first setting (corresponding to tensor $\bbA$ displayed in the first row of Figure~\ref{fig:ground_truth}). 
We observe that these weights are far from being uniform, and effectively induce a strongly varying scaling across kernels $k=1, 2, 3$ and between nodes.
Although this display is hard to interpret, it can be better understood when looked together with the first row of Figure~\ref{fig:ground_truth}: we observe a similarly looking block structure, which means that these weights scale the penalization level roughly following the block structure of the adjacency matrix $\bbA$ and the intensity of the baseline vector $\mu$.

\section{Conclusion}

In this paper we proposed a careful analysis of the generalization
error of the multivariate Hawkes process.
Our theoretical analysis required a new concentration
inequality for matrix-martingales in continuous time, with an observable variance term, 
which is a result of independent interest. 
This analysis led to a new data-driven tuning
of sparsity-inducing penalizations, that we assessed on a numerical example.
Future works will focus on other theoretical results for non-convex matrix factorization
techniques applied to this problem.

\section{Proofs} %
\label{sec:proofs}

This Section contains the proofs of all the results given in the paper.
First, we prove the statements concerned with deviation inequalities, namely Theorems~\ref{thm:concentration_counting_th},~\ref{thm:concentration_counting_emp}, Proposition~\ref{lem1} and Theorem~\ref{thm:concentration_counting_emp1}.
Then, we give the proof of Theorem~\ref{thm:fast-oracle}, concerning the oracle inequality for the procedure.

\subsection{Proof of Theorem \ref{thm:concentration_counting_th}}
\label{app:conc1}

In~\cite{bacry_gaiffas_muzy_1}, a deviation inequality is proven in a slightly more general setting than the one considered in this paper.
There are mainly two differences.
\begin{itemize}
\item This paper considers only counting processes with uniform jumps of size 1 whereas in~\cite{bacry_gaiffas_muzy_1}, jump sizes are controlled by a predictable process $\bJ$.
Therefore, it suffices to set $\bJ = {\textbf 1}$ and $\bC_s = {\textbf 1}$ in Equations~(2) and~(3) of~\cite{bacry_gaiffas_muzy_1}, where ${\textbf 1}$ stands for the all-ones matrices with relevant shapes.

\item In~\cite{bacry_gaiffas_muzy_1}, the deviation inequality is proved in a general context where no symmetry is assumed on~$\tT_s$. It forces to consider a symmetric version of  $\bW_{\tT}(s)$ as in Eq.~\eqref{eq:W_def1} increasing the dimension of the working space by a factor of 2, which leads to less precise deviation inequality. In this paper we consider both cases, symmetric and non symmetric, in order to obtain slightly better constants (see the definition of $K_{m,n}$).
\end{itemize}
With those two differences in mind, following carefully the proof of the concentration inequality in
\cite{bacry_gaiffas_muzy_1} (see the beginning of Appendix B.1 herein) one gets
  \begin{align*}
    \P \bigg[ \frac{\lmax(\sS(\bZ_t))}{b} \geq \frac 1 \xi \lmax
    &\Big(\int_0^t \frac{\phi\big(\xi J_{\max} \norm{\bC_s}_\infty
    \max(\norm{\tT_s}_{\op; \infty},\norm{\tT_s^\top}_{\op; \infty}) b^{-1}\big)}{J_{\max}^2 \norm{\bC_s}_\infty^2  \max(\norm{\tT_s}^2_{\op; \infty},\norm{\tT_s^\top}^2_{\op; \infty})}
    \bW_s ds \Big) + \frac x \xi, \\
    & b_{\tT}(t) \le b \bigg] \leq (m + n) e^{-x},
  \end{align*}
where $\xi \in (0,3)$ and $\lmax(\sS(\bZ_t)) = \normop{\bZ}$ (see the beginning of Appendix
B.1 in~\cite{bacry_gaiffas_muzy_1}).
Setting $\bJ = {\textbf 1}$, $\bC = {\textbf 1}$ and taking care of the symmetric case at the same time as the non symmetric one, one gets:
  \begin{align*}
    \P \bigg[ \frac{\normop{\bZ_t}}{b} \geq \frac 1 \xi \lmax
    \Big( &\int_0^t \frac{\phi\big(\xi
    \max(\norm{\tT_s}_{\op; \infty},\norm{\tT_s^\top}_{\op; \infty}) b^{-1}\big)}{ \max(\norm{\tT_s}^2_{\op; \infty},\norm{\tT_s^\top}^2_{\op; \infty})}
    \bW_s ds \Big) + \frac x \xi, \\
    & b_{\tT}(t) \le b \bigg] \leq K_{m,n} e^{-x},
  \end{align*}
  using the definitions $K_{m,n}$ and $\bW_s$ introduced previously (depending on the symmetric properties of the tensor $\tT_s$).
	Let us note that on $\{ b_{\tT}(t) \leq b \}$ one has $  \max(\norm{\tT_s}_{\op; \infty},\norm{\tT_s^\top}_{\op; \infty})b^{-1} \leq 1$ for any $s \in [0, t]$.
Thus, since  $\phi(x h)\leq h^2 \phi(x)$ for any $h \in [0,1]$ and $x
    > 0$, one gets
  \begin{equation*}
    \P \bigg[ \frac{\normop{\bZ_t}}{b} \geq \frac {\phi(\xi)} {\xi b^2} \lmax
    \Big(\int_0^t
    \bW_s ds \Big) + \frac x \xi~,~~~b_{\tT}(t) \le b \bigg] \leq K_{m,n} e^{-x}
  \end{equation*}
  and finally
  \begin{equation*}
    \P \bigg[ \normop{\bZ_t} \geq \frac {\phi(\xi)} {\xi b} \lmax
  (\bV_t) + \frac {xb} \xi~,~~~b_{\tT}(t) \le b \bigg] \leq K_{m,n} e^{-x}
  \end{equation*}
 which proves the first part of the Theorem.
 The second part (i.e., Inequality \eqref{eqcor1:1}) can be obtained following
 some standard tricks (see e.g.~\cite{massart2007concentration}):
 \begin{itemize}
  \item[(i)] on $(0,3)$, $\phi(\xi) \leq \frac{\xi^2}{2(1-\xi/3)}$ and
  \item[(ii)]   $\min_{\xi \in (0, 1/c)} \big( \frac{a \xi}{1 - c \xi}
    + \frac x \xi \big) = 2 \sqrt{ax} + c x$ for any $a, c, x > 0$.
\end{itemize}
Thus applying (i) leads to
  \begin{equation*}
    \P \bigg[ \normop{\bZ_t} \geq \frac{\xi}{2b(1-\xi/3)}  \lmax
  (\bV_t) + \frac {xb} \xi~,~~~b_{\tT}(t) \le b \bigg] \leq K_{m,n} e^{-x}
  \end{equation*}
  or equivalently
  \begin{equation*}
    \P \bigg[ \normop{\bZ_t} \geq \frac{\xi}{2b(1-\xi/3)}v + \frac {xb} \xi~,~~~\lmax
  (\bV_t)\le v,~b_{\tT}(t) \le b \bigg] \leq K_{m,n} e^{-x}.
  \end{equation*}
  Then optimizing on $\xi$ using (ii) with $c=1/3$ and $a=v/2b^2$, one gets
\begin{equation*}
    \P \bigg[ \normop{\bZ_t} \geq
\sqrt{2vx } + \frac{xb} 3~,~~~\lmax
  (\bV_t)\le v,~b_{\tT}(t) \le b \bigg] \leq K_{m,n} e^{-x}
  \end{equation*}
  which concludes the proof of Theorem~\ref{thm:concentration_counting_th}.

\subsection{Proof of Proposition~\ref{lem1}}
\label{app:lemm1}

This Proposition provides a deviation between $\lambda_{\max}(\bV(t))$ and $\lambda_{\max}(\widehat \bV(t))$.
Let us notice that it is a generalization to arbitrary matrices of dimensions $m \times n$ of an analog
inequality originally proven by~\cite{hansen_reynaud_bouret_viroirard} for scalar martingales (i.e., in dimension 1). The proof below follows the same lines as these authors.
The proof is based on the observation that the difference $\bV_\tT(t) - \hbV_\tT(t)$ can be written as a martingale $\bZ_\tH(t)$
\begin{equation*}
	\bV_\tT(t) - \hbV_\tT(t) = \bZ_\tH(t) = \int_0^t \tH_s \circ  d \bM_s,
\end{equation*}
where
\begin{equation}
\label{eq:H_def2}
\tH_s = \tT_s^2
\end{equation}
when $\tT_s$ is symmetric, while
\begin{equation}
\label{eq:H_def1}
\tH_s =
\begin{bmatrix}
\tT_s \tT_s^\top & \bO \\ \bO &
\tT_s^\top \tT_s
\end{bmatrix}
\end{equation}
if $\tT_s$ is not symmetric.
Then applying Eq. \eqref{eqthm1:1} of Theorem \ref{thm:concentration_counting_th} to the martingale $\bZ_\tH(t)$ (we are in the symmetric case of the Theorem since
$\tH^\top_s = \tH_s$), one gets
\begin{equation}
\P \bigg[ \normop{\bZ_\tH(t)} \geq  \frac{\phi(\xi)}{\xi b} \lmax \big(\bV_\tH(t))  + \frac{x b}{\xi}, \quad b_\tH(t) \leq b \bigg] \leq K_{m,n} e^{-x},
\end{equation}
with
\begin{equation}
\label{defvth}
 \bV_{\tH}(t) = \int_0^t \tH_s^2 \circ \blambda_s ds \; .
\end{equation}
Since
\begin{equation*}
	\normop{\bZ_\tH(t)} \geq \lambda_{\max}(\bV_\tT(t)) - \lambda_{\max}(\hbV_\tT(t)),
\end{equation*}
we have
\begin{equation}
\label{ineq1}
\P \bigg[ \lambda_{\max}(\bV_{\tT}(t)) \geq \lambda_{\max}(\widehat \bV_{\tT}(t))+ \frac{\phi(\xi)}{\xi b} \lmax \big(\bV_\tH(t))  + \frac{x b}{\xi}, \quad b_\tH(t) \leq b \bigg] \leq K_{m,n} e^{-x},
\end{equation}
One can first notice that, from the definitions of $\tH$ and $b_\tT(t)$, one has $b_\tH(t) \leq b^2_\tT(t)$.
Moreover, since
\begin{equation*}
\tT_s \tT_s^\top \mleq b_\tT^2(s) \bI_m \quad \mbox{and} \quad \tT_s^\top \tT_s \mleq b^2_\tT(s) \bI_n
\end{equation*}
for all $s$, we have from Eq.~\eqref{defvth},
\begin{equation}
  \bV_\tH(t) \mleq b^2_\tT(t) \bV_\tT(t)
\end{equation}
and therefore
\begin{equation*}
	\lmax(\bV_\tH(t)) \leq b^2_\tT(t) \lmax(\bV_\tT(t)).
\end{equation*}
Inequality \eqref{ineq1} then gives:
\begin{equation}
\P \bigg[ \lambda_{\max}(\bV_{\tT}(t)) \geq \lambda_{\max}(\widehat \bV_{\tT}(t))+ \frac{\phi(\xi)}{\xi}  \lambda_{\max}(\bV_{\tT}(t))  + \frac{x b^2}{\xi}, \quad b_\tT(t) \leq b \bigg] \leq K_{m,n} e^{-x},
\end{equation}
and thus
\begin{equation}
\P \bigg[ \lambda_{\max}(\bV_{\tT}(t)) \geq \frac{\xi \lambda_{\max}(\widehat \bV_{\tT}(t))}{\xi-\phi(\xi)}  + \frac{x b^2}{\xi-\phi(\xi)}, \quad b_\tT(t) \leq b \bigg] \leq K_{m,n} e^{-x},
\end{equation}
which proves the first inequality stated in Proposition~\ref{lem1}.
Now, an easy computation proves that the choice $\xi = -W_{-1}(-\frac{2}{3} e^{-2/3})-2/3 \approx 0.762$
provides the second desired inequality. $\hfill \square$

\subsection{Proof of Theorem~\ref{thm:concentration_counting_emp}} % (fold)
\label{sub:proof_of_theorem_thm:concentration_counting_emp}

Introduce the set
\begin{equation*}
	E_t = \{ \lambda_{\max}(\bV_\tT(t)) \leq 2 \lambda_{\max}(\hbV_\tT(t)) + 2.62 b^2 x \}.
\end{equation*}
We know from Proposition~\ref{lem1} that $\P[E_t^\complement, b_\tT(t) \leq b] \leq K_{m, n} e^{-x}$.
Now, on the set
\begin{equation*}
	E_t \cap \{ \lambda_{\max}(\hbV_\tT(t)) \leq v \} \cap \{ b_\tT(t) \leq b \}
\end{equation*}
we have
\begin{align*}
	\frac{\phi(\xi)}{\xi b} \lmax(\bV(t)) + \frac{x b}{\xi} \leq \frac{\phi(\xi)}{\xi b} 2 v
	+ \frac{b x}{\xi} + \frac{2.62 \phi(3)}{3} b x
\end{align*}
for any $\xi \in (0, 3)$, since $\xi \mapsto \phi(\xi) / \xi$ is increasing. Using again points~(i) and~(ii) from Section~\ref{app:conc1} proves that the minimum for $\xi \in (0, 3)$ of the right hand size of this last inequality is equal to
\begin{equation*}
	2 \sqrt{v x} + \frac{2.62 \phi(3) + 1}{3} x b \leq 2 \sqrt{v x} + c x b
\end{equation*}
with $c = 14.39$.
Now, the conclusion easily follows from the following decomposition:
\begin{align*}
	\P \bigg[ &\normop{\bZ_\tT(t)} \geq 2 \sqrt{v x} + c b x, \;\;
	\lambda_{\max}(\hbV_\tT(t)) \leq v, \; \; b_\tT(t) \leq b \bigg] \\
	&\leq \P[E_t^\complement, b_\tT(t) \leq b] + \P \bigg[ \normop{\bZ_\tT(t)} \geq 2 \sqrt{v x} + c b x,
	\;\; E_t, \;\; \lambda_{\max}(\hbV_\tT(t)) \leq v, \; \; b_\tT(t) \leq b \bigg] \\
	&\leq K_{m, n} e^{-x} + \P \bigg[ \normop{\bZ_t} \geq \frac{\xi}{2b(1-\xi/3)}  \lmax
	  (\bV_t) + \frac {xb} \xi~,~~~b_{\tT}(t) \le b \bigg] \\
    &\leq 2 K_{m, n} e^{-x},
\end{align*}
where we used Equation~\eqref{eqthm1:1} from Theorem~\ref{thm:concentration_counting_th} in
the last inequality.

\subsection{Proof of Theorem \ref{thm:concentration_counting_emp1}}
\label{app:th3}

In order to prove this theorem, we are going to use peeling arguments. For any $\epsilon > 0$ and $z > 0$ we define the interval
\begin{equation*}
	{\cal I}_{z,\varepsilon} = [z,z(1+\varepsilon)].
\end{equation*}
Let, $v_0, b_0, \epsilon >0$ and let us define
$v_j = v_0 (1+\varepsilon)^j$, $b_j = b_0 (1+\varepsilon)^j$.
Let us define also the events
\begin{equation*}
	V_{-1} = \{  \lmax(\widehat \bV_{\tT}(t)) \leq v_0 \}, \quad
	 B_{-1}= \{ b_\tT(t) \leq b_0 \},
\end{equation*}
and
\begin{equation*}
	V_j = \{ \lmax(\widehat \bV_{\tT}(t)) \in {\cal I}_{v_j,\varepsilon} \}, \quad B_j = \{ b_\tT(t) \in {\cal I}_{b_j,\varepsilon} \}
\end{equation*}
for any $j \in \N$.
We set $v_0 = w_0 x$, then, from Equation~\eqref{eqcor2:1}, one gets successively
\begin{align*}
   \P \bigg[ \normop{\bZ_\tT(t)} \geq x \big(2 \sqrt{w_0}  +
  c b_0 \big), V_{-1} \cap B_{-1} \bigg]  &\leq 2 K_{m,n} e^{-x} \\
  \P \bigg[ \normop{\bZ_\tT(t)} \geq x \big(2 \sqrt{w_0}  +
 c (1+\varepsilon) b_\tT(t) \big), V_{-1} \cap B_{j} \bigg]
 & \leq 2 K_{m,n} e^{-x} \\
   \P \bigg[ \normop{\bZ_\tT(t)} \geq 2
   \sqrt{\lmax(\widehat \bV_{\tT}(t)) (1+\varepsilon) x} +
 c x b_0, V_{i} \cap B_{-1} \bigg]  & \leq 2 K_{m,n} e^{-x} \\
   \P \bigg[ \normop{\bZ_\tT(t)} \geq 2
   \sqrt{\lmax(\widehat \bV_{\tT}(t)) (1+\varepsilon) x} +
   c (1+\varepsilon) x b_\tT(t), V_{i} \cap B_{j} \bigg]
   & \leq 2 K_{m,n} e^{-x}
\end{align*}
for all $i,j \geq 0$.
If one denotes $A = 2 \sqrt{w_0} / c + b_0$, previous inequalities entail, for any $i,j \geq -1$:
\begin{equation}
\label{eq:A}
\P \bigg[ \normop{\bZ_\tT(t)} \geq 2
\sqrt{\lmax(\widehat \bV_{\tT}(t)) (1 + \varepsilon) x} +
c (1+\varepsilon) (A + b_\tT(t)) x, V_{i} \cap B_{j} \bigg]
\leq  2 K_{m,n} e^{-x}.
\end{equation}
Let
$\alpha >0$ and define
\begin{equation}
	\ell_x(t) = \alpha \log \bigg( \log \Big(
	\frac{ \lmax(\widehat \bV_{\tT}(t))}{w_0 x}(1+\epsilon)^2 \vee (1+\epsilon) \Big) \bigg) + \alpha \log \bigg( \log \Big(\frac{b_\tT(t)}{b_0} (1+\epsilon)^2 \vee (1+\epsilon) \Big) \bigg).
\end{equation}
Since, $\forall i,j\ge -1$, $ \lmax(\widehat \bV_{\tT}(t)) \geq xw_0 (1+\varepsilon)^{i}(1-\delta_{-1,i})$ and $b_\tT(t) \geq b_0(1+\varepsilon)^{j}(1-\delta_{-1,j})$
on $V_i \cap B_j$, then one has
\begin{equation*}
	\ell_x(t) \ge \ell_{i,j} = \log \Big( (i+2)^\alpha (j+2)^\alpha
	(\log(1 + \epsilon))^{2\alpha} \Big) \quad \text{on} \quad V_i \cap B_j
\end{equation*}
for any $i,j \ge -1$.
Then making the change of variable $x \leftarrow x + \ell_{i,j}$
in~\eqref{eq:A} gives
\begin{align*}
	\P \bigg[ \normop{\bZ_\tT(t)} \geq
	2 & \sqrt{\lmax(\widehat \bV_{\tT}(t)) (1 + \varepsilon)
	(x + \ell_{i,j})} + c(1 + \varepsilon)(A + b_\tT(t)) (x + \ell_{i,j}),
	 \;\; V_{i} \cap B_{j} \bigg] \\
	 & \leq 2 K_{m,n} e^{-x} e^{-\ell_{i,j}}
\end{align*}
and then
\begin{align*}
\P \bigg[ \normop{\bZ_\tT(t)} \geq 2 &\sqrt{\lmax(\widehat \bV_{\tT}(t))
(1 + \varepsilon) (x+\ell_x(t))} + c (1+\varepsilon) (x + \ell_x(t))
(A + b_\tT(t)), \quad V_{i} \cap B_{j} \bigg]  \\
& \leq  2 K_{m,n} \big[\log(1 + \varepsilon)\big]^{-2\alpha}
e^{-x} \big[(i+2)(j+2)\big]^{-\alpha}
\end{align*}
for any $i,j \ge -1$.
Since the whole probability space can be partitioned as
$\bigcup_{i,j \in \geq -1} V_i \cap B_j$, one has finally
\begin{align*}
  \P \bigg[ &\normop{\bZ_\tT(t)} \geq 2 \sqrt{\lmax(\widehat \bV_{\tT}(t))
  (1 + \varepsilon) (x + \ell_x(t))} + c(1 + \varepsilon) (x+\ell_x(t)) (A + b_\tT(t)) \bigg] \\
  & =\sum_{i,j=-1}^{\infty} \P \bigg[ \normop{\bZ_\tT(t)} \geq
  2 \sqrt{\lmax(\widehat \bV_{\tT}(t)) (1 + \varepsilon) (x + \ell_x(t))} \\
  & \quad \quad \quad \quad \quad \quad + c(1 + \varepsilon) (x+\ell_x(t)) (A+b_\tT(t)), \quad V_{i} \cap B_{j} \bigg] \\
  & \leq 2 K_{m,n} \big[\log(1+\varepsilon)\big]^{-2\alpha}
   \big(\sum_{i=1}^\infty i^{-\alpha} \big)^2 e^{-x}.
\end{align*}
Finally, choosing $\epsilon = b_0 = w_0 = 1$ and $\alpha = 2$ leads to Equation~\eqref{eqth3:1} and concludes the proof of the Theorem.

\subsection{Proof of Theorem~\ref{thm:fast-oracle}}
\label{sec:proof_of_oracles}

If $A, B$ are vectors, matrices or tensors of matching dimensions, we denote by $A \odot B$ their entrywise product (Hadamard product).
We recall also that $\bA_{j, \bul}$ the $j$-th row of a matrix $\bA$ and recall that $\norm{\bA}_{\infty, 2} = \max_j \norm{\bA_{j, \bul}}_2$.
The proof is based on the proof of a sharp oracle inequality for trace
norm penalization, see~\cite{koltchinskii2011nuclear} and~\cite{koltchinskii2011oracle}. 
We endow the space $\R^d \times \R^{d \times d \times K}$ with the inner product
\begin{equation*}
  \inr{\theta, \theta'} = \inr{\mu, \mu'} + \inr{\bbA, \bbA'},
\end{equation*}
where $\theta = (\mu, \bbA)$ and
$\theta' = (\mu', \bbA')$ with $\inr{\mu, \mu'} = \mu^\top \mu'$ and
\begin{equation*}
  \inr{\bbA, \bbA'} = \sum_{\substack{1 \leq j, j' \leq d \\ 1 \leq k \leq K}}
  \bbA_{j, j', k} \bbA_{j, j', k}'.
\end{equation*}
We denote for short $a_{j, j', k} = \bbA_{j, j', k}$.
For any $\theta$, one has
\begin{equation*}
  \inr{\grad R_T(\hat \theta), \hat \theta - \theta} = 2 \sum_{1 \leq
    j \leq d} (\hat \mu_j - \mu_j) \frac{\partial R_T( \hat
    \theta)}{\partial \hat \mu_j} + \sum_{\substack{1 \leq j, j' \leq d \\ 1 \leq k \leq K}} (\hat
  a_{j, j', k} - a_{j, j', k}) \frac{\partial R_T(\hat \theta)}{\partial \hat a_{j, j', k}}.
\end{equation*}
Let us recall that $H_{j, j', k}(t) = \int_{(0, t)} h_{j, j', k}(t - s) d N_{j'}(s)$.
Since
\begin{equation*}
  \frac{\partial \lambda_{j, \theta}(t)}{\partial \mu_j} = 1 \quad
  \text{ and } \quad \frac{\partial \lambda_{j, \theta}(t)}{\partial
    a_{j, j', k}} = H_{j, j', k}(t),
\end{equation*}
we have that the derivatives of the empirical risk are given by
\begin{equation*}
  \frac{\partial R_T( \hat \theta)}{\partial \mu_j} = \frac 2T
  \Big( \int_0^T \lambda_{j, \theta}(t) dt - \int_0^T d N_j(t) \Big)
\end{equation*}
and
\begin{align*}
  \frac{\partial R_T( \hat \theta)}{\partial a_{j, j', k}} 
  &= \frac 2T \Big( \int_0^T H_{j, j', k}(t) \lambda_{j, \theta}(t) dt 
  - \int_0^T H_{j, j', k}(t) d N_j(t) \Big).
\end{align*}
It leads to
\begin{align*}
  \inr{\grad R_T(\hat \theta), \hat \theta - \theta} &= \frac 2T
  \sum_{j=1}^d \int_0^T (\lambda_{j, \hat \theta}(t) - d N_j(t) )
  (\hat \mu_j - \mu_j) \\
  & \quad + \frac 2T\sum_{\substack{1 \leq j, j' \leq d \\ 1 \leq k \leq K}} 
  \int_0^T H_{j, j', k}(t) 
  (\lambda_{j, \hat \theta}(t) - d N_j(t) )
  (\hat a_{j, j', k} - a_{j, j', k}) \\
  &= \frac 2T \sum_{j=1}^d \int_0^T (\lambda_{j, \hat \theta}(t) -
  \lambda_{j, \theta}(t)) (\lambda_{j, \hat \theta}(t) dt - d N_j(t)).
\end{align*}
Let us remind that $M_j(t) = N_j(t) - \int_0^t \lambda_j(s) ds$ are martingales coming from the Doob-Meyer decomposition, so that $d M_j(t) = d N_j(t) - \lambda_j(t) dt$.
So, recalling that
\begin{equation*}
  \inr{f, g}_T = \frac 1T \sum_{1 \leq j \leq d} \int_{[0, T]} f_j(t)
  g_j(t) dt,
\end{equation*}
we obtain the decomposition
\begin{align*}
  \inr{\grad R_T(\hat \theta), \hat \theta - \theta} &= 2
  \inr{\lambda_{\hat \theta} - \lambda_{\theta}, \lambda_{\hat \theta}
    - \lambda}_T - \frac 2T \sum_{j=1}^d \int_0^T (\lambda_{j, \hat
    \theta}(t) - \lambda_{j, \theta}(t)) d M_j(t).
\end{align*}
Namely, we end up with
\begin{align}
\label{eq:1}
  2 \inr{\lambda_{\hat \theta} - \lambda_{\theta}, \lambda_{\hat
      \theta} - \lambda}_T = \inr{\grad R_T(\hat \theta), \hat
    \theta - \theta} + \frac 2T \sum_{j=1}^d \int_0^T (\lambda_{j,
    \hat \theta}(t) - \lambda_{j, \theta}(t)) d M_j(t).
\end{align}
The parallelogram identity gives
\begin{equation*}
  2 \inr{\lambda_{\hat \theta} - \lambda_{\theta},
    \lambda_{\hat \theta} - \lambda}_T = \norm{\lambda_{\hat \theta} -
    \lambda}_T^2 + \norm{\lambda_{\hat \theta} - \lambda_{\theta}}_T^2 -
  \norm{\lambda_{\theta} - \lambda}_T^2,
\end{equation*}
where we put $\norm{f}_{T}^2 = \inr{f, f}_T$.
Let us point out that, in the case $ \inr{\lambda_{\hat \theta} -
  \lambda_{\theta}, \lambda_{\hat \theta} - \lambda}_T < 0$, one
obtains
\begin{equation*}
  \norm{\lambda_{\hat \theta} - \lambda}_T^2 \leq
  \norm{\lambda_{\theta} - \lambda}_T^2,
\end{equation*}
which directly implies the inequality of the Theorem.
Thus, from now on, let us assume that
\begin{equation}
\label{eq:2}
  \inr{\lambda_{\hat \theta} - \lambda_{\theta}, \lambda_{\hat
      \theta} - \lambda}_T \geq 0.
\end{equation}
The first order condition for $ \hat \theta \in \argmin_{\theta} \{
R_T(\theta) + \pen(\theta) \}$ gives
\begin{equation*}
\label{eq:subgrad}
-\grad R_T(\hat \theta) \in \partial \pen(\hat
\theta).
\end{equation*}
Let $\hat \theta_\partial  = -\grad R_T(\hat \theta)$.
Since the subdifferential is a monotone mapping, we have $\inr{\hat
  \theta - \theta, \hat \theta_\partial - \theta_\partial} \geq 0$ for
any
$\theta_\partial \in \partial \pen(\theta)$. Thus from \eqref{eq:1}, one gets
$\forall \theta_\partial \in \partial \pen(\theta)$,
\begin{align}
\label{eq:3}
  2 \inr{\lambda_{\hat \theta} - \lambda_{\theta}, \lambda_{\hat
      \theta} - \lambda}_T \leq -\inr{\theta_{\partial}, \hat
    \theta - \theta} + \frac 2T \sum_{j=1}^d \int_0^T (\lambda_{j,
    \hat \theta}(t) - \lambda_{j, \theta}(t)) d M_j(t).
\end{align}
We need now to
characterize the structure of the subdifferentials involved in
$\pen(\theta)$, to describe $\theta_\partial$.
If $g_1(\mu) = \sum_{j = 1}^d \hat w_{j} |\mu_j|$, for $\hat w_{j} \geq 0$, we have
\begin{equation}
  \label{eq:subdiff_g1}
  \partial g_1(\mu) = \Big\{ \hat w \odot \sign(\mu) + \hat w \odot
  f : \norm{f}_\infty \leq 1, \mu \odot f = 0 \Big\}.
\end{equation}
If $g_2(\bbA) = \sum_{1 \leq j, j' \leq d, 1 \leq k \leq K} \hat \bbW_{j, j', k} 
|\bbA_{j, j', k}|$, for $\hat \bbW_{j, j', k} \geq 0$, we have
\begin{equation}
  \label{eq:subdiff_g2}
  \partial g_2(\bbA) = \Big\{ \hat \bbW \odot \sign(\bbA) + \hat \bbW \odot
  \bb F : \norm{\bb F}_\infty \leq 1, \bb A \odot \bb F = 0 \Big\}.
\end{equation}
Now let $\bs A = \hstack(\bb A)$ and $\hat {\bs A} = \hstack(\hat {\bb A})$.
Let us recall that if $\bs A = \bs U \bs \Sigma \bs V^\top$ is the SVD
of $\bs A$, we have $\cP_\bA(\bB) = \bP_\bU \bB + \bB \bP_\bV -
\bP_\bU \bB \bP_\bV$ and $\cP_{\bs A}^\perp(\bs B) = (\bI - \bP_\bU)
\bB (\bI - \bP_\bV)$ (projection onto the column and row space of
$\bA$ and projection onto its orthogonal space). Now, for $g_3(\bs A)
= \hat \tau \norm{\bs A}_*$, we have
\begin{equation}
  \label{eq:subdiff_g3}
  \partial g_3(\bs A) = \Big\{ \hat \tau \bs U \bs V^\top + \hat \tau \cP_{\bs
    A}^\perp(\bs F) : \normop{\bs F} \leq 1 \Big\},
\end{equation}
see for instance~\citep{lewis1995convex}. 
Now, write
\begin{equation*}
  -\inr{\theta_{\partial}, \hat \theta - \theta} =
  -\inr{\mu_{\partial}, \hat \mu - \mu} -\inr{\bbA_{\partial, 1}, \hat
    \bbA - \bbA} -\inr{\bA_{\partial, *}, \hat \bA - \bA}
\end{equation*}
with $\mu_{\partial} \in \partial g_1(\mu)$, $\bbA_{\partial, 1} \in \partial  g_2(\bbA)$
and $\bA_{\partial, *} \in \partial  g_3(\bA)$. 
Using Equation~\eqref{eq:subdiff_g1},~\eqref{eq:subdiff_g2}
and~\eqref{eq:subdiff_g3}, we can write
\begin{align*}
  -\inr{\theta_{\partial}, \hat \theta - \theta} &= -\inr{\hat w \odot
    \sign(\mu), \hat \mu - \mu} -\inr{\hat w \odot f,
    \hat \mu - \mu} \\
  &\quad \quad - \inr{\hat \bbW \odot \sign(\bb A), \hat \bbA - \bbA}
  -\inr{\hat \bbW \odot \bb F_1, \hat \bbA - \bbA} \\
  &\quad \quad - \hat \tau \inr{\bU \bV^\top, \hat \bA - \bA} - \hat
  \tau \inr{\bF_*, \cP_{\bA}^\perp(\hat \bA - \bA)},
\end{align*}
where by duality between the norms $\norm{\cdot}_1$ and
$\norm{\cdot}_\infty$, and between $\norm{\cdot}_*$ and
$\normop{\cdot}$, we can choose $f, \bb F_1$ and $\bF_*$ such that
\begin{equation*}
  \inr{\hat w \odot f, \hat \mu - \mu} = \norm{(\hat \mu -
    \mu)_{\supp(\mu)^\perp}}_{1, \hat w}, \quad \inr{\hat {\bb W} \odot
    \bb F_1, \hat {\bb A} - \bb A} = \norm{(\hat \bbA - \bbA)_{\supp(\bbA)^\perp}}_{1, \hat \bbW}
\end{equation*}
and
\begin{equation*}
  \inr{\bF_*, \cP_{\bA}^\perp(\hat \bA - \bA)} =
  \norm{\cP_{\bA}^\perp(\hat \bA - \bA)}_*,
\end{equation*}
which leads to
\begin{align*}
  -\inr{\theta_{\partial}, \hat \theta - \theta} &\leq \norm{(\hat \mu
    - \mu)_{\supp(\mu)}}_{1, \hat w} - \norm{(\hat \mu -
    \mu)_{\supp(\mu)^\perp}}_{1, \hat w} \\
  &\quad \quad + \norm{(\hat \bbA - \bbA)_{\supp(\bbA)}}_{1, \hat \bbW} -
  \norm{(\hat \bbA - \bbA)_{\supp(\bbA)^\perp}}_{1, \hat \bbW} \\
  &\quad \quad + \hat \tau \norm{\cP_{\bA}(\hat \bA - \bA)}_* - \hat
  \tau \norm{\cP_{\bA}^\perp(\hat \bA - \bA)}_*.
\end{align*}
Now, we decompose the noise term of~\eqref{eq:3}:
\begin{align*}
  \frac 2T &\sum_{j=1}^d \int_0^T (\lambda_{j, \hat \theta}(t) -
  \lambda_{j, \theta}(t)) d M_j(t) \\
  &= \frac 2T \sum_{j=1}^d (\hat \mu_j - \mu_j) 
  \int_0^T d M_j(t) + \frac 2T \sum_{\substack{1 \leq j, j' \leq d \\1 \leq k \leq K}}
  (\hat a_{j, j', k} - a_{j, j', k}) \int_0^T H_{j, j', k}(t) d M_{j}(t) \\
  &= \frac 2T \inr{\hat \mu - \mu, M(T)} + \frac 2T \inr{\hat \bbA - \bbA, \bbZ(T)},
\end{align*}
where $M(T) = [M_1(T) \cdots M_d(T)]^\top$ and where $\bbZ(T)$ is the 
$d \times d \times K$ tensor with entries
\begin{equation*}
  \bbZ_{j, j', k}(T) = \int_0^T H_{j, j', k}(t) d M_{j}(t).
\end{equation*}
Recall that $\hstack$ is the horizontally stacking operator defined by~\eqref{eq:hstack_definition}.
The following upper bounds
\begin{align*}
  |\inr{\hat \mu - \mu, M(T)}| &\leq \sum_{j=1}^d |\hat \mu_j -
  \mu_j| |M_j(T)| \\
  |\inr{\hat \bbA - \bbA, \bbZ(T)} | &\leq 
  \sum_{\substack{1 \leq j, j' \leq d \\1 \leq k \leq K}} 
  |\hat \bbA_{j, j', k} - \bbA_{j, j', k}| |\bbZ_{j, j', k}(T)| \\
  |\inr{\hat \bbA - \bbA, \bbZ(T)} | &= \inr{\hstack(\hat \bbA - \bbA), \hstack(\bbZ(T))} 
  \leq \normop{\hstack(\bbZ(T))} \norm{\hstack(\hat \bbA - \bbA)}_*,
\end{align*}
entail that we need to upper bound the three terms
\begin{equation*}
  |M_j(T)|, \quad |\bbZ_{j, j', k}(T)| \quad \text{ and } 
  \quad \normop{\hstack(\bbZ(T))}
\end{equation*}
by data-driven quantities.
Let us start with $\normop{\hstack(\bbZ(T))}$. Denote for short $\bZ(t) = \hstack(\bbZ(t))$ and  $\bH(t) = \hstack(\bbH(t))$ where $\bbH(t)$ is defined by~\eqref{eq:def_bbH}.
We note that
\begin{equation*}
  \bZ(t) = \int_0^t \diag(d M(s)) \bH(s),
\end{equation*}
namely
\begin{equation*}
  (\bZ(t))_{j, j' + (k-1) d} = \int_0^t (\bbH(t - s))_{j, j', k} d M_j(s)
\end{equation*}
for any $1 \leq j, j' \leq d$ and $1 \leq k \leq K$.
We need the following corollary.

\begin{corollary}
  \label{cor:deviation-particular-case}

 The following deviation inequality holds
\begin{equation}
  \label{eq:deviation-particular-case}
  \begin{split}
    \P \bigg[ \normop{\bZ(t)} \geq 2 &\sqrt{\lambda_{\max}(\hbV(t)) (x + \log(2d) + 
    \ell(t))} \\
    & \quad \quad \quad + 14.39 (x + \log(2d) + \ell(t)) (1 + \sup_{0 \leq s \leq t} 
    \norm{\bH(s)}_{\infty, 2}) \bigg] \leq 23.45 e^{-x},    
  \end{split}
\end{equation}
where
\begin{equation*}
  \lmax(\widehat \bV(t)) = \lmax\Big(\int_0^t \bH^\top(s) \bH(s) \diag(d N(s))\Big)
  \; \bigvee \; \max_{j=1, \ldots, d} \int_0^t \norm{\bH_{j, \bul}(s)}_2^2 d N_j(s),
\end{equation*}
and where
\begin{equation*}
  \ell(t) = 2 \log \log \Big (\frac{4 \lambda_{\max}(\hbV(t))}{x}
  \vee 2  \Big) + 2 \log \log \Big(4  \sup_{0 \leq s \leq t}  
  \norm{\bH(s)}_{\infty, 2} \vee 2 \Big).
\end{equation*}
\end{corollary}

The proof of Corollary~\ref{cor:deviation-particular-case} is given in Section~\ref{sub:proof_of_corollary} below.
Corollary~\ref{cor:deviation-particular-case} proves that
$\frac 1T \normop{\bZ(t)} \leq \frac{\hat \tau}{2}$ holds with probability $1 - 23.45 e^{-x}$, with
\begin{align*}
  \hat \tau &= 4 \sqrt{\frac{\lambda_{\max}(\hbV(T) / T) (x + \log(2d) + 
  \ell(T))}{T}} \\
    & \quad \quad + 28.78 \frac{x + \log(2d) + \ell(T)) 
  (1 + \sup_{0 \leq t \leq T} \norm{\bH(t)}_{\infty, 2})}{T},
\end{align*}
which leads to the choice of $\hat \tau$ given in Section~\ref{sec:procedure}.
This entails that, on an event of probability larger than $1 - 23.45 e^{-x}$, we have
\begin{equation*}
  \frac 1T | \inr{\hat \bbA - \bbA, \bbZ(T)} | \leq \frac{\hat \tau}{2} 
  \norm{\hstack(\hat \bbA - \bbA)}_*.
\end{equation*}
Using again Corollary~\ref{cor:deviation-particular-case} with $\bH(t) \equiv 1$ (constant number equal to 1) and $M = M_j$ gives that $\frac 1T |M_j(T)| \leq \frac{\hat w_j}{3}$
 for all $j=1, \ldots, d$ with probability $1 - 23.45 e^{-x}$ with
\begin{equation*}
  \hat w_j = 6 \sqrt{\frac{(N_j(T) / T) (x + \log d + 
  \ell_j(T))}{T}} + 86.34 \frac{x + \log d + \ell_j(T)}{T},
\end{equation*}
with $\ell_j(T) = 2 \log \log ( \frac{4 N_j(T)}{x} \vee 2) + 2 \log \log 4$.
This entails that, on an event of probability larger than $1 - 23.45 e^{-x}$, we have
\begin{equation*}
  \frac 2T |\inr{\hat \mu - \mu, M(T)}| \leq \frac 23 \norm{\hat \mu - \mu}_{1, \hat w}.
\end{equation*}
Using a last time Corollary~\ref{cor:deviation-particular-case} with $\bH(t) = H_{j,j', k}(t)$ 
and $M = M_j$ gives $\frac 1T |\bbZ_{j, j', k}(T)| \leq \frac{\hat \bbW_{j,j', k}}{2}$ 
uniformly for $j, j', k$ for
\begin{align*}
  \hat \bbW_{j,j', k} &= 4 \sqrt{\frac{\frac 1T \int_0^T H_{j, j', k}(t)^2 d N_j(t) (x + \log(K d^2) 
  + \bb L_{j, j', k}(T))}{T}} \\
  & \quad + 28.78\frac{(x + \log(K d^2) + \bb L_{j,j', k}(T))(1 + \sup_{0 \leq t \leq T} |H_{j, j', k}(t)|)}{T},
\end{align*}
where 
\begin{equation*}
  \bb L_{j,j', k}(T) = 2 \log \log \Big( \frac{4 \int_0^T H_{j, j', k}(t)^2 d N_j(t)}{x} \vee 2 \Big)
   + 2 \log \log \Big(4 \sup_{0 \leq t \leq T} |H_{j, j', k}(t)| \vee 2 \Big),
\end{equation*}
which entails that on an event of probability larger than $1 - 23.45 e^{-x}$, we have
\begin{equation*}
  \frac 1T |\inr{\hat \bbA - \bbA, \bbZ(T)}| \leq \frac 12 \norm{\hat \bbA - \bbA}_{1, \hat \bbW}.
\end{equation*}
This entails that, with a probability larger than $1 - 3 \times 23.45 e^{-x}$, one has
\begin{align*}
  0 \leq -\inr{\theta_{\partial}, \hat \theta - \theta} &+ \frac 2T
  \sum_{j=1}^d \int_0^T (\lambda_{j, \hat \theta}(t) - \lambda_{j,
    \theta}(t)) d M_j(t) \\
  &\leq \frac 53 \norm{(\hat \mu - \mu)_{\supp(\mu)}}_{1, \hat w} -
  \frac 13 \norm{(\hat \mu - \mu)_{\supp(\mu)^\perp}}_{1, \hat w} \\
  &\quad + \frac 32 \norm{(\hat {\bbA} - {\bbA})_{\supp({\bbA})}}_{1, \hat \bbW}
  - \frac 12 \norm{(\hat {\bbA} - {\bbA})_{\supp({\bbA})^\perp}}_{1, \hat \bbW}
  \\ &\quad + \frac 32 \hat \tau \norm{\cP_{\bA}(\hat \bA - \bA)}_* -
  \frac 12 \hat \tau \norm{\cP_{\bA}^\perp(\hat \bA - \bA)}_*,
\end{align*}
where we recall once again that $\bA = \hstack(\bbA)$ and $\hat \bA = \hstack(\hat \bbA)$.
This matches the constraint of Definition~\ref{ass:RE} with $\mu' = \hat \mu - \mu$ and 
$\bbA' = \hat \bbA - \bbA$, so that it entails
% Taking $\bA = \hat \bA$ gives a cone constraint on $\hat \mu - \mu$:
% \begin{equation*}
%   \norm{(\hat \mu - \mu)_{\supp(\mu)^\perp}}_{1, \hat w} \leq 5 \norm{(\hat
%     \mu - \mu)_{\supp(\mu)}}_{1, \hat w},
% \end{equation*}
% while taking $\mu = \hat \mu$ gives a cone constraint on $\hat \bbA -
% \bbA$:
% \begin{align*}
%   \norm{(\hat \bbA - \bbA)_{\supp(\bbA)^\perp}}_{1, \hat \bbW} & + \hat
%   \tau \norm{\cP_{\bA}^\perp(\hat \bA - \bA)}_* \\
%   &\leq 3 \norm{(\hat \bbA - \bbA)_{\supp(\bbA)}}_{1, \hat \bbW} + 3 \hat
%   \tau \norm{\cP_{\bA}(\hat \bA - \bA)}_*.
% \end{align*} 
\begin{equation}
  \label{eq:using_RE}
  \norm{(\hat \mu - \mu)_{\supp(\mu)}}_{2} \vee \norm{(\hat \bbA -
    \bbA)_{\supp(\bbA)}}_{F} \vee \norm{\cP_\bA(\hat \bA - \bA)}_{F}
  \leq \kappa(\theta) \norm{\lambda_{\hat \theta} -
    \lambda_{\theta}}_T.
\end{equation}
Putting all this together gives
\begin{align*}
  - \inr{\theta_\partial, &\hat \theta - \theta} + \frac 2T \inr{\hat
    \mu - \mu, M(T)} + \frac 2T \inr{\hat \bbA - \bbA,
    \bbZ(T)} \\
  &\leq \frac 53 \norm{(\hat \mu - \mu)_{\supp(\mu)}}_{1, \hat w} -
  \frac 13 \norm{(\hat \mu - \mu)_{\supp(\mu)^\perp}}_{1, \hat w} \\
  &\quad + \frac 32 \norm{(\hat \bbA - \bbA)_{\supp(\bbA)}}_{1, \hat \bbW}
  - \frac 12 \norm{(\hat \bbA -
    \bbA)_{\supp(\bbA)^\perp}}_{1, \hat \bbW} \\
  &\quad + \frac 32 \hat \tau \norm{\cP_{\bA}(\hat \bA - \bA)}_* -
  \frac 12 \hat \tau \norm{\cP_{\bA}^\perp(\hat \bA - \bA)}_* \\
  &\leq \frac 53 \norm{(\hat w)_{\supp(\mu)}}_2 \norm{(\hat \mu -
    \mu)_{\supp(\mu)}}_{2} + \frac 32 \norm{(\hat \bbW)_{\supp(\bbA)}}_F
  \norm{(\hat \bbA - \bbA)_{\supp(\bbA)}}_F \\
  &\quad + \frac 32 \hat \tau \sqrt{\rank(\bA)} \norm{\cP_{\bA}(\hat
    \bA - \bA)}_F,
\end{align*}
where we used Cauchy-Schwarz's inequality. This finally gives
\begin{align*}
  \norm{\lambda_{\hat \theta} - \lambda}_T^2 &\leq
  \norm{\lambda_{\theta} - \lambda}_T^2 - \norm{\lambda_{\hat \theta}
    - \lambda_{\theta}}_T^2 \\
  & \quad + \kappa(\theta) \Big(\frac 53 \norm{(\hat
    w)_{\supp(\mu)}}_2 + \frac 32 \norm{(\hat \bbW)_{\supp(\bbA)}}_F +
  \frac 32 \hat \tau \sqrt{\rank(\bA)} \Big) \norm{\lambda_{\hat
      \theta} - \lambda_{\theta}}_T
\end{align*}
where we used~\eqref{eq:using_RE}. The conclusion of the proof of
Theorem \ref{thm:fast-oracle} follows from the fact that $ax - x^2
\leq a^2 / 4$ for any $a, x > 0$.

\subsection{Proof of Corollary~\ref{cor:deviation-particular-case}} % (fold)
\label{sub:proof_of_corollary}

We simply use Theorem~\ref{thm:concentration_counting_emp1}.
First, we remark that $\bZ(t) = \int_0^t \tT(s) \circ \diag(d M(s))$ for the tensor $\tT(t)$ of size $d \times K d \times d \times d$ given by
\begin{equation}
  \label{eq:particular-tensor}
	(\tT(t))_{i, j; k, l} = (\bI)_{i, k} (\bH(t))_{l, j}
\end{equation}
for $1 \leq i, k, l \leq d$ and $1 \leq j \leq Kd$.
Note that we have
\begin{equation}
  \label{eq:explicit-tensor-submatrices}
	\tT_{\bul, \bul; k, l}(t) = e_k \bH_{l, \bul}(t)^\top \quad \text{ and } \quad
	\tT_{\bul, \bul; k, l}(t)^\top = \bH_{l, \bul}(t) e_k^\top
\end{equation}
where $e_k \in \R^d$ stands for the $k$-th element of the canonical basis of $\R^d$
and where $\bH_{l, \bul}(t) \in \R^{Kd}$ stands for the vector corresponding to the $l$-th row of the matrix $\bH(t)$.
Therefore, we have
\begin{equation*}
	\tT_{\bul, \bul; k, l}(t) \tT_{\bul, \bul; k, l}^\top(t) =
	\norm{\bH_{l, \bul}(t)}_2^2 e_k e_k^\top \quad \text{and} \quad
	\tT_{\bul, \bul; k, l}^\top(t) \tT_{\bul, \bul; k, l}(t) =
	\bH_{l, \bul}(t) \bH_{l, \bul}(t)^\top
\end{equation*}
and therefore
\begin{equation*}
	\normop{\tT_{\bul, \bul; k, l}(t)} = \sqrt{\lmax(\tT_{\bul, \bul; k, l}(t)
	\tT_{\bul, \bul; k, l}^\top(t))} = \norm{\bH_{l, \bul}(t)}_2
\end{equation*}
and
\begin{equation*}
	\norm{\tT(t)}_{\op; \infty}	= \max_{1 \leq l \leq d} \norm{\bH_{l, \bul}(t)}_2 
  = \norm{\bH(t)}_{\infty, 2}.
\end{equation*}
One can prove in the same way that $\norm{\tT^\top(t)}_{\op; \infty} = 
\norm{\bH(t)}_{\infty, 2}$, so that for this choice
of tensor $\tT(t)$, we have $b_{\tT}(t) = \norm{\bH(t)}_{\infty, 2}$.
Now, let us explicit what $\widehat \bV_{\tT}(t)$ is for the tensor~\eqref{eq:particular-tensor}.
First, let us remind that
\begin{equation*}
\hbV_\tT(t) =
\begin{bmatrix}
\int_0^t \tT(s) \tT^\top(s) \circ \diag(d N(s)) & \bO \\ \bO &
\int_0^t \tT^\top(s) \tT(s) \circ \diag(d N(s))
\end{bmatrix}.
\end{equation*}
Using~\eqref{eq:explicit-tensor-submatrices} we get
\begin{equation*}
  (\tT(t) \tT(t)^\top)_{\bul, \bul;, k, l} = e_k \bH_{l, \bul}(t)^\top \bH_{l, \bul}(t)
   e_k^\top= \norm{\bH_{l, \bul}(t)}_2^2 e_k e_k^\top
\end{equation*}
so that $\int_0^t (\tT(s) \tT^\top(s)) \circ \diag(d N(s))$ is the diagonal matrix with entries
\begin{equation*}
  \Big(\int_0^t (\tT(s) \tT^\top(s)) \circ \diag(d N(s)) \Big)_{j, j} =
  \int_0^t \norm{\bH_{j, \bul}(s)}_2^2 d N_j(s),
\end{equation*}
or equivalently
\begin{equation*}
  \int_0^t (\tT(s) \tT^\top(s)) \circ \diag(d N(s)) =
  \int_0^t \diag(\bH^\top(s) \bH(s)) \diag(d N(s)).
\end{equation*}
Using again~\eqref{eq:explicit-tensor-submatrices} we get
\begin{equation*}
  (\tT^\top(t) \tT(t))_{\bul, \bul;, k, l} = \bH_{l, \bul}(t) e_k^\top e_k
  \bH_{l, \bul}(t)^\top = \bH_{l, \bul}(t) \bH_{l, \bul}(t)^\top
\end{equation*}
so that $\int_0^t (\tT^\top(s) \tT(s)) \circ \diag(d N(s))$ is the matrix with entries
\begin{equation*}
  \Big(\int_0^t (\tT^\top(s) \tT(s)) \circ \diag(d N(s)) \Big)_{i, j} =
  \sum_{l=1}^d \int_0^t \bH_{l, i}(s) \bH_{l, j}(s) d N_l(s)
\end{equation*}
or equivalently
\begin{equation*}
  \int_0^t (\tT^\top(s) \tT(s)) \circ \diag(d N(s)) =
  \int_0^t \bH^\top(s) \bH(s) \diag(d N(s)).
\end{equation*}
Finally, we obtain that
\begin{equation*}
  \lmax(\widehat \bV_t) = \lmax\Big(\int_0^t \bH^\top(s) \bH(s) \diag(d N(s))\Big)
  \; \bigvee \; \max_{j=1, \ldots, d} \int_0^t \norm{\bH_{j, \bul}(t)}_2^2 d N_j(s).
\end{equation*}
This concludes the proof of the corollary. $\hfill \square$

\acks{
This work was funded in part by the French government under management of Agence Nationale de la Recherche as part of the "Investissements d'avenir" program, reference ANR-19-P3IA-0001 (PRAIRIE 3IA Institute).
}


\begin{thebibliography}{40}
\providecommand{\natexlab}[1]{#1}
\providecommand{\url}[1]{\texttt{#1}}
\expandafter\ifx\csname urlstyle\endcsname\relax
  \providecommand{\doi}[1]{doi: #1}\else
  \providecommand{\doi}{doi: \begingroup \urlstyle{rm}\Url}\fi

\bibitem[Bacry et~al.(2013)Bacry, Delattre, Hoffmann, and
  Muzy]{bacry2013modelling}
E.~Bacry, S.~Delattre, M.~Hoffmann, and J.-F. Muzy.
\newblock Modelling microstructure noise with mutually exciting point
  processes.
\newblock \emph{Quantitative Finance}, 13\penalty0 (1):\penalty0 65--77, 2013.

\bibitem[Bacry et~al.(2015)Bacry, Mastromatteo, and Muzy]{bacryHawkesFinance}
E.~Bacry, I.~Mastromatteo, and J.-F. Muzy.
\newblock Hawkes processes in finance.
\newblock \emph{Market Microstructure and Liquidity}, 01\penalty0
  (01):\penalty0 1550005, 2015.

\bibitem[Bacry et~al.(2016{\natexlab{a}})Bacry, Ga{\"\i}ffas, Mastromatteo, and
  Muzy]{bacry2016mean}
E.~Bacry, S.~Ga{\"\i}ffas, I.~Mastromatteo, and J.-F. Muzy.
\newblock Mean-field inference of hawkes point processes.
\newblock \emph{Journal of Physics A: Mathematical and Theoretical},
  49\penalty0 (17):\penalty0 174006, 2016{\natexlab{a}}.

\bibitem[Bacry et~al.(2016{\natexlab{b}})Bacry, Ga\"iffas, and
  Muzy]{bacry_gaiffas_muzy_1}
E.~Bacry, S.~Ga\"iffas, and J.-F. Muzy.
\newblock Concentration inequalities for matrix martingales in continuous time.
\newblock \emph{Probability Theory and Related Fields}, 170:\penalty0 525--553,
  2016{\natexlab{b}}.

\bibitem[Bacry et~al.(2018)Bacry, Bompaire, Deegan, Ga\"{\i}ffas, and
  Poulsen]{tick}
E.~Bacry, M.~Bompaire, P.~Deegan, S.~Ga\"{\i}ffas, and S.~V. Poulsen.
\newblock tick: a python library for statistical learning, with an emphasis on
  hawkes processes and time-dependent models.
\newblock \emph{Journal of Machine Learning Research}, 18\penalty0
  (214):\penalty0 1--5, 2018.

\bibitem[Bartlett and Mendelson(2006)]{bartlett2006empirical}
P.~L. Bartlett and S.~Mendelson.
\newblock Empirical minimization.
\newblock \emph{Probability Theory and Related Fields}, 135\penalty0
  (3):\penalty0 311--334, 2006.

\bibitem[Beck and Teboulle(2009)]{fista}
A.~Beck and M.~Teboulle.
\newblock A fast iterative shrinkage-thresholding algorithm for linear inverse
  problems.
\newblock \emph{SIAM Journal of Imaging Sciences}, 2\penalty0 (1):\penalty0
  183--202, 2009.

\bibitem[Bickel et~al.(2009)Bickel, Ritov, and Tsybakov]{MR2533469}
P.~J. Bickel, Y.~Ritov, and A.~B. Tsybakov.
\newblock Simultaneous analysis of lasso and {D}antzig selector.
\newblock \emph{Ann. Statist.}, 37\penalty0 (4):\penalty0 1705--1732, 2009.

\bibitem[Blundell et~al.(2012)Blundell, Heller, and
  Beck]{blundell2012modelling}
C.~Blundell, K.~A Heller, and J.~M. Beck.
\newblock Modelling reciprocating relationships with hawkes processes.
\newblock In \emph{NIPS}, pages 2609--2617, 2012.

\bibitem[Bompaire(2018)]{bompaire_phd}
M.~Bompaire.
\newblock \emph{Machine Learning based on Hawkes processes and Stochastic
  Optimization}.
\newblock PhD thesis, CMAP, Ecole polytechique, EDMH, 2018.

\bibitem[Bompaire et~al.(2018)Bompaire, Bacry, and
  Ga{\"\i}ffas]{bompaire2018dual}
M.~Bompaire, E.~Bacry, and S.~Ga{\"\i}ffas.
\newblock Dual optimization for convex constrained objectives without the
  gradient-lipschitz assumption.
\newblock \emph{arXiv preprint arXiv:1807.03545}, 2018.

\bibitem[Cand{\`e}s and Tao(2004)]{Candes04}
E.~J. Cand{\`e}s and T.~Tao.
\newblock Decoding by linear programming.
\newblock \emph{IEEE Transactions on Information Theory}, 12\penalty0
  (51):\penalty0 4203--4215, 2004.

\bibitem[Cand{\`e}s and Tao(2009)]{Candes09}
E.~J. Cand{\`e}s and T.~Tao.
\newblock The power of convex relaxation: Near-optimal matrix completion.
\newblock \emph{IEEE Transactions on Information Theory}, 56\penalty0 (5),
  2009.

\bibitem[Crane and Sornette(2008)]{crane2008robust}
R.~Crane and D.~Sornette.
\newblock Robust dynamic classes revealed by measuring the response function of
  a social system.
\newblock \emph{Proceedings of the National Academy of Sciences}, 105\penalty0
  (41), 2008.

\bibitem[Daneshmand et~al.(2014)Daneshmand, Rodriguez, Song, and
  Sch\"{o}lkpof]{gomez14a}
N.~Daneshmand, M.~Rodriguez, L.~Song, and B.~Sch\"{o}lkpof.
\newblock Estimating diffusion network structure: Recovery conditions, sample
  complexity, and a soft-thresholding algorithm.
\newblock \emph{ICML}, 2014.

\bibitem[de~Menezes and Barab\'asi(2004)]{PhysRevLett.92.028701}
M.~Argollo de~Menezes and A.-L. Barab\'asi.
\newblock Fluctuations in network dynamics.
\newblock \emph{Phys. Rev. Lett.}, 92:\penalty0 028701, Jan 2004.
\newblock \doi{10.1103/PhysRevLett.92.028701}.
\newblock URL \url{http://link.aps.org/doi/10.1103/PhysRevLett.92.028701}.

\bibitem[DuBois et~al.(2013)DuBois, Butts, and Smyth]{dubois2013stochastic}
C.~DuBois, C.~Butts, and P.~Smyth.
\newblock Stochastic blockmodeling of relational event dynamics.
\newblock In \emph{Proceedings of the Sixteenth International Conference on
  Artificial Intelligence and Statistics}, pages 238--246, 2013.

\bibitem[Gomez-Rodriguez et~al.(2013)Gomez-Rodriguez, Leskovec, and
  Sch\"olkopf]{gomez13}
M.~Gomez-Rodriguez, J.~Leskovec, and B.~Sch\"olkopf.
\newblock Modeling information propagation with survival theory.
\newblock \emph{ICML}, 2013.

\bibitem[Hansen et~al.(2012)Hansen, Reynaud-Bouret, and
  Rivoirard]{hansen_reynaud_bouret_viroirard}
N.~R. Hansen, P.~Reynaud-Bouret, and V.~Rivoirard.
\newblock Lasso and probabilistic inequalities for multivariate point
  processes.
\newblock Technical report, Arvix preprint, 2012.

\bibitem[Hawkes(1971)]{hawkes1971spectra}
A.~G. Hawkes.
\newblock Spectra of some self-exciting and mutually exciting point processes.
\newblock \emph{Biometrika}, 58\penalty0 (1):\penalty0 83--90, 1971.

\bibitem[Iwata et~al.(2013)Iwata, Shah, and Ghahramani]{iwata2013discovering}
T.~Iwata, A.~Shah, and Z.~Ghahramani.
\newblock Discovering latent influence in online social activities via shared
  cascade poisson processes.
\newblock In \emph{Proceedings of the 19th ACM SIGKDD international conference
  on Knowledge discovery and data mining}, pages 266--274. ACM, 2013.

\bibitem[Koltchinskii(2011)]{koltchinskii2011oracle}
V.~Koltchinskii.
\newblock \emph{Oracle Inequalities in Empirical Risk Minimization and Sparse
  Recovery Problems: Saint-Flour XXXVIII-2008}, volume 2033.
\newblock Springer, 2011.

\bibitem[Koltchinskii et~al.(2011)Koltchinskii, Lounici, and
  Tsybakov]{koltchinskii2011nuclear}
V.~Koltchinskii, K.~Lounici, and A.~B. Tsybakov.
\newblock Nuclear-norm penalization and optimal rates for noisy low-rank matrix
  completion.
\newblock \emph{The Annals of Statistics}, 39\penalty0 (5):\penalty0
  2302--2329, 2011.

\bibitem[Leskovec(2008)]{Leskovec08}
J.~Leskovec.
\newblock \emph{Dynamics of large networks}.
\newblock PhD thesis, Machine Learning Department, Carnegie Mellon University,
  2008.

\bibitem[Leskovec et~al.(2009)Leskovec, Backstrom, and
  Kleinberg]{leskovec2009meme}
J.~Leskovec, L.~Backstrom, and J.~Kleinberg.
\newblock Meme-tracking and the dynamics of the news cycle.
\newblock In \emph{Proceedings of the 15th ACM SIGKDD}. ACM, 2009.

\bibitem[Lewis(1995)]{lewis1995convex}
A.~S. Lewis.
\newblock The convex analysis of unitarily invariant matrix functions.
\newblock \emph{Journal of Convex Analysis}, 2\penalty0 (1):\penalty0 173--183,
  1995.

\bibitem[Linderman and Adams(2014)]{linderman2014discovering}
S.~W. Linderman and R.~P. Adams.
\newblock Discovering latent network structure in point process data.
\newblock \emph{arXiv preprint arXiv:1402.0914}, 2014.

\bibitem[Massart(2007)]{massart2007concentration}
P.~Massart.
\newblock \emph{Concentration inequalities and model selection}, volume 1896.
\newblock Springer, 2007.

\bibitem[Mohler et~al.(2011)Mohler, Short, Brantingham, Schoenberg, and
  Tita]{mohler2011self}
G.~O. Mohler, M.~B. Short, P.~J. Brantingham, F.~P. Schoenberg, and G.~E. Tita.
\newblock Self-exciting point process modeling of crime.
\newblock \emph{Journal of the American Statistical Association}, 2011.

\bibitem[Ogata(1978)]{ogata1978asymptotic}
Y.~Ogata.
\newblock The asymptotic behaviour of maximum likelihood estimators for
  stationary point processes.
\newblock \emph{Annals of the Institute of Statistical Mathematics},
  30\penalty0 (1):\penalty0 243--261, 1978.

\bibitem[Ogata(1981)]{ogata1981lewis}
Y.~Ogata.
\newblock On lewis' simulation method for point processes.
\newblock \emph{Information Theory, IEEE Transactions on}, 27\penalty0
  (1):\penalty0 23--31, 1981.

\bibitem[Ogata(1998)]{ogata1998space}
Y.~Ogata.
\newblock Space-time point-process models for earthquake occurrences.
\newblock \emph{Annals of the Institute of Statistical Mathematics},
  50\penalty0 (2):\penalty0 379--402, 1998.

\bibitem[Pino et~al.(1999)Pino, Landesa, Rodriguez, Obelleiro, and
  Burkholder]{pino1999generalized}
M.~R. Pino, L.~Landesa, J.~L. Rodriguez, F.~Obelleiro, and R.~J. Burkholder.
\newblock The generalized forward-backward method for analyzing the scattering
  from targets on ocean-like rough surfaces.
\newblock \emph{IEEE Transactions on Antennas and Propagation}, 47\penalty0
  (6):\penalty0 961--969, 1999.

\bibitem[Ricci et~al.(2011)Ricci, Rokach, and Shapira]{ricci2011introduction}
F.~Ricci, L.~Rokach, and B.~Shapira.
\newblock \emph{Introduction to recommender systems handbook}.
\newblock Springer, 2011.

\bibitem[Richard et~al.(2014)Richard, Ga\"iffas, and Vayatis]{richard14}
E.~Richard, S.~Ga\"iffas, and N.~Vayatis.
\newblock Link prediction in graphs with autoregressive features.
\newblock \emph{Journal of Machine Learning Research}, 2014.

\bibitem[Rodriguez et~al.(2011)Rodriguez, Balduzzi, and
  Sch{\"o}lkopf]{rodriguez2011uncovering}
M.~Rodriguez, D.~Balduzzi, and B.~Sch{\"o}lkopf.
\newblock Uncovering the temporal dynamics of diffusion networks.
\newblock \emph{ICML}, 2011.

\bibitem[Tropp(2012)]{tropp2012user}
J.~A. Tropp.
\newblock User-friendly tail bounds for sums of random matrices.
\newblock \emph{Foundations of Computational Mathematics}, 12\penalty0
  (4):\penalty0 389--434, 2012.

\bibitem[Van De~Geer(2000)]{van2000empirical}
S.~Van De~Geer.
\newblock \emph{Empirical Processes in M-estimation}, volume 105.
\newblock Cambridge university press Cambridge, 2000.

\bibitem[Yang and Zha(2013)]{yang2013mixture}
S.-H. Yang and H.~Zha.
\newblock Mixture of mutually exciting processes for viral diffusion.
\newblock In \emph{ICML}, 2013.

\bibitem[Zhou et~al.(2013)Zhou, Zha, and Song]{zhou2013learning}
K.~Zhou, H.~Zha, and L.~Song.
\newblock Learning social infectivity in sparse low-rank networks using
  multi-dimensional hawkes processes.
\newblock In \emph{AISTATS}, volume~31, pages 641--649, 2013.

\end{thebibliography}
\end{document}